\documentclass[preprint]{elsarticle}

\usepackage[toc,page]{appendix}
\usepackage[utf8]{inputenc}
\usepackage{float} 
\usepackage{hyperref}
\usepackage{caption}
\usepackage{subcaption}
\usepackage{amssymb}
\usepackage{amsmath}
\usepackage{amsthm}
\usepackage{nccmath}
\usepackage{graphicx}
\usepackage{color}
\usepackage{multirow}
\usepackage{mathtools}
\usepackage{fixltx2e}
\usepackage{algpseudocode}
\usepackage{algorithm}
\usepackage{cancel}
\usepackage{colortbl}
\usepackage{textcomp}
\usepackage{tikz}

\journal{Artificial Intelligence}

\MakeRobust{\Call}

\makeatletter
\renewcommand{\ALG@beginalgorithmic}{\small}
\makeatother

\hypersetup{colorlinks=false,pdfborder={0 0 0},
bookmarksopen,bookmarksdepth=3}

\newtheorem{defassumption}{Default Assumption}

\theoremstyle{plain}

\newtheorem{theorem}{Theorem}

\theoremstyle{definition}
\newtheorem{definition}{Definition}
\newtheorem{example}{Example}[section]

\newcommand{\crel}{
\begin{tikzpicture}[baseline=0.5ex]%
\draw (0,0) -- (0,2ex);%
\draw (0,1ex) -- (1ex,1.5ex);%
\draw (0,1ex) -- (1ex,0.5ex);%
\end{tikzpicture}%
}

\newcommand{\ncrel}{
\begin{tikzpicture}[baseline=0.5ex]%
\draw (0,0) -- (0,2ex);%
\draw (0,1ex) -- (1ex,1.5ex);%
\draw (0,1ex) -- (1ex,0.5ex);%
\draw (-0.4ex, 0ex) -- (1.4ex, 2ex);%
\end{tikzpicture}%
}

\title{On the adoption of abductive reasoning for time series interpretation}
\author{T. Teijeiro}
\ead{tomas.teijeiro@usc.es}
\author{P. Félix}
\address{Centro Singular de Investigación en Tecnoloxías da Información 
(CITIUS), University of Santiago de Compostela, Santiago de Compostela, Spain}

\begin{document}

\begin{abstract}
Time series interpretation aims to provide an explanation of what is observed in 
terms of its underlying processes. The present work is based on the assumption 
that the common classification-based approaches to time series interpretation 
suffer from a set of inherent weaknesses, whose ultimate cause lies in the 
monotonic nature of the deductive reasoning paradigm. In this document we 
propose a new approach to this problem, based on the initial hypothesis that 
abductive reasoning properly accounts for the human ability to identify and 
characterize the patterns appearing in a time series. The result of this 
interpretation is a set of conjectures in the form of observations, organized 
into an abstraction hierarchy and explaining what has been observed. A 
knowledge-based framework and a set of algorithms for the interpretation task 
are provided, implementing a hypothesize-and-test cycle guided by an attentional 
mechanism. As a representative application domain, interpretation of the 
electrocardiogram allows us to highlight the strengths of the proposed approach 
in comparison with traditional classification-based approaches.
\end{abstract}


\maketitle


\section{Introduction}

The interpretation and understanding of the behavior of a complex system 
involves the deployment of a cognitive apparatus aimed at guessing the processes 
and mechanisms underlying what is observed. The human ability to recognize 
patterns plays a paramount role as an instrument for highlighting evidence which 
should require an explanation, by matching information from observations with 
information retrieved from memory. Classification naturally arises as a pattern 
recognition task, defined as the assignment of observations to categories.

Let us first state precisely at this point what is the problem under 
consideration: we wish to interpret the behavior of a complex system by 
measuring a physical quantity along time. This quantity is represented as a time 
series.

The Artificial Intelligence community has devoted a great deal of effort on 
different paradigms, strategies, methodologies and techniques for time series 
classification. Nonetheless, in spite of the wide range of proposals for 
building classifiers, either by eliciting domain knowledge or by induction from 
a set of observations, the resulting classifiers behave as deductive systems. 
The present work is premised on the assumption that some of the important 
weaknesses of this approach lie in its deductive nature, and that an abductive 
approach can address these shortcomings, which are described below.

Let us remember that a deduction contains in its conclusions information that is 
already implicitly contained in the premises, and thus it is truth-preser-ving. 
In this sense, a classifier ultimately assigns a label or a set of labels to 
observations. This label can designate a process or a mechanism of the system 
being observed, but it is nothing more than a term that summarizes the premises 
implied by the observations. Conversely, abduction, or inference to the best 
explanation, is a form of inference that goes from data to a hypothesis that 
best explains or accounts for the data~\cite{Peirce31}. Abductive conclusions 
contain new information not contained in the premises, and are capable of 
predicting new evidence, although they are fallible. Abductions are thus 
truth-widening, and they can make the leap from the language of observations to 
the language of the underlying processes and mechanisms, responding to the 
aforementioned problem in a natural way \cite{Josephson94}. For example, 
consider a simple rule stating that if a patient experiences a sudden 
tachycardia and a decrease in blood pressure, then we can conclude that he or 
she is suffering from shock due to a loss of blood volume. From a deductive 
perspective, \textit{loss of blood volume} is just a name provided by the rule 
for the satisfaction of the two premises. However, from an abductive 
perspective, \textit{loss of blood volume} is an explanatory hypothesis, a 
conjecture, that expands the truth contained in the premises, enabling the 
observer to predict additional consequences such as, for example, pallid skin, 
faintness, dizziness or thirst.

Of course, the result of a classifier can be considered as a conjecture, but 
always from an external agent, since a classifier is monotonic as a logical 
system and its conclusions cannot be refuted from within. Classifier ensembles 
aim to overcome the errors of individual classifiers by combining different 
classification instances to obtain a better result; thus, a classifier can be 
amended by others in the final result of the ensemble. However, even an ensemble 
represents a bottom-up mapping, and classification invariably fails above a 
certain level of distortion within the data. The interpretation and 
understanding of a complex system usually unfolds along a set of abstraction 
layers, where at each layer the temporal granularity of the representation is 
reduced from below. A classification strategy provides an interpretation as the 
result of connecting a set of classifiers along the abstraction structure, and 
the monotonicity of deduction entails a propagation of errors from the first 
abstraction layers upwards, narrowing the capability of making a proper 
interpretation as new abstraction layers are successively added. Following an 
abductive process instead, an observation is conjectured at each abstraction 
layer as the best explanatory hypothesis for the data from the layer or layers 
below, within the context of information from above, and the non-monotonicity of 
abduction supports the retraction of any observation at any abstraction layer in 
the search for the best global explanation. Thus, bottom-up and top-down 
processing complement one another and provide a joint result. As a consequence, 
abduction can guess the underlying processes from corrupted data or even in the 
temporary absence of data. 

On the other hand, a classifier is based on the assumption that the underlying 
processes or mechanisms are mutually exclusive. Superpositions of two or more 
processes  are excluded; they must be represented by a new process, 
corresponding to a new category which is different and usually unrelated to 
previous ones. Therefore, an artificial casuistry-based heuristics is adopted, 
increasing the complexity of the interpretation and reducing its adaptability to 
the variability of observations. In contrast, abduction can reach a conclusion 
from the availability of partial evidence, refining the result by the 
incremental addition of new information. This makes it possible to discern 
different processes just from certain distinguishable features, and at the end 
to infer a set of explanations as far as the available evidence does not allow 
us to identify the best one, and they are not incompatible with each other.

In a classifier, the truth of the conclusion follows from the truth of all the 
premises, and missing data usually demand an imputation strategy that results in 
a conjecture: a sort of abducing to go on deducing. In contrast, an abductive 
interpretation is posed as a hypothesize-and-test cycle, in which missing data 
are naturally managed, since a hypothesis can be evoked by every single piece of 
evidence in isolation and these can be incrementally added to reasoning. This 
fundamental property of abduction is well suited to the time-varying 
requirements of the interpretation of time series, where future data can compel 
changes to previous conclusions, and the interpretation task may be requested to 
provide the current result as the best explanation at any given time.

Abduction has primarily been proposed for diagnostic tasks~\cite{Console91a, 
Peng90}, but also for question answering \cite{Ferrucci2012}, language 
understanding \cite{Hobbs93}, story comprehension \cite{Charniak88}, image 
understanding \cite{Poole90} or plan recognition \cite{Litman87}, amongst 
others. Some studies have proposed that perception might rely on some form of 
abduction. Even though abductive reasoning has been proven to be NP-complete or 
worse, a compiled form of abduction based on a set of pre-stored hypotheses 
could narrow the generation of hypotheses \cite{Josephson94}. The present work 
takes this assumption as a starting point and proposes a model-based abductive 
framework for time series interpretation supported on a set of temporal 
abstraction patterns. An abstraction pattern represents a set of constraints 
that must be satisfied by some evidence in order to be interpreted as the 
hypothetical observation of a certain process, together with an observation 
procedure providing a set of measurements for the features of the conjectured 
observation. A set of algorithms is devised in order to achieve the best 
explanation through a process of successive abstraction from raw data, by means 
of a hypothesize-and-test strategy.

Some previous proposals have adopted a non-monotonic schema for time series 
interpretation. TrenDx system detects significant trends in time series by 
matching data to predefined trend patterns \cite{Haimowitz93, Haimowitz95}. One 
of these patterns plays the role of the expected or normal pattern, and the 
other patterns are fault patterns. A matching score of each pattern is based on 
the error between the pattern and the data. Multiple trend patterns can be 
maintained as competing hypotheses according to their matching score; as 
additional data arrive some of the patterns can be discarded and new patterns 
can be triggered. This proposal has been applied to diagnose pediatric growth 
trends. A similar proposal can be found in \cite{Larizza95}, taking a step 
further by providing complex temporal abstractions, the result of finding out 
specific temporal relationships between a set of significant 
trends. This proposal has been applied to the infectious surveillance of heart 
transplanted patients. Another example is the Résumé system, a knowledge-based 
temporal abstraction framework \cite{Shahar96,Shahar97}. Its goal is to provide 
a set of interval-based temporal abstractions from time-stamped input data, 
distinguishing four output abstraction types: state, gradient, rate and pattern. 
It uses a truth maintenance system to retract inferred intervals that are no 
longer true, and propagate new abstractions. Furthermore, this framework 
includes a non-monotonic interpolation mechanism for trend detection 
\cite{Shahar99}. This approach has been applied to several clinical domains 
(protocol-based care, monitoring of children's growth and therapy of diabetes) 
and to an engineering domain (monitoring of traffic control). 

The present work includes several examples and results from the domain of 
electrocardiography. The electrocardiogram (ECG) is the recording at the body's 
surface of the electrical activity of the heart as it changes with time, and is 
the primary method for the study and diagnosis of cardiac disease, since the 
processes involved in cardiac physiology manifest in characteristic temporal 
patterns on the ECG trace. In other words, a correct reading of the ECG has the 
potential to provide valuable insight into cardiac phenomena. Learning to 
interpret the ECG involves the acquisition of perceptual skills from an 
extensive bibliography with interpretation criteria and worked examples. In 
particular, pattern recognition is especially important in order to build a 
bottom-up representation of cardiac phenomena in multiple abstraction levels. 
This has encouraged extensive research on classification techniques for 
interpreting the ECG; however, in spite of all these efforts, this is still 
considered an open problem. We shall try to demonstrate that the problem lies in 
the nature of deduction itself.

The rest of this paper is structured as follows: Section~\ref{sec:motivexample} 
introduces the main concepts and terminology used in the paper in an informal 
and intuitive way. Following this, in Sections~\ref{sec:definitions}, 
\ref{sec:framework} and \ref{sec:search} we formally describe all the components 
of the interpretation framework, including the knowledge representation model 
and the algorithms used to obtain effective interpretations within an affordable 
time. Section~\ref{sec:strengths} illustrates the capabilities of the framework 
in overcoming some of the most important shortcomings of deductive classifiers. 
Section~\ref{sec:evaluation} presents the main experimental results derived from 
this work. Finally, in section~\ref{sec:discussion} we discuss the properties of 
the model compared with other related approaches and draw several conclusions.

\section{Interpretation as a process-guessing task}
\label{sec:motivexample}
We propose a knowledge-based interpretation framework upon the principles of 
abductive reasoning, on the basis of a strategy of hypothesis formation and 
testing. Taking as a starting point a time series of physical measurements, a 
set of observations are guessed as conjectures of the underlying processes, 
through successive levels of abstraction. Each new observation will be generated 
from previous levels as the underlying processes aggregate, superimpose or 
concatenate to form more complex processes with greater duration and scope, and 
are organized into an abstraction hierarchy. 

The knowledge of the domain is described as a set of abstraction patterns as
follows:

\begin{eqnarray*}
[h_\psi(\mathbf{A}_h,T^b_h,T^e_h)=\varTheta(\mathbf{A}_1,T_1,..., \mathbf{A}_n , 
T_n)] 
& abstracts ~ m_1(\mathbf{A}_1,T_1),..., m_n(\mathbf{A}_n,T_n) \\
&\quad \{C(\mathbf{A}_h,T^b_h,T^e_h,\mathbf{A}_1,T_1,...,\mathbf{A}_n,T_n)\}
\end{eqnarray*}

\noindent where $h_\psi(\mathbf{A}_h,T^b_h, T^e_h)$ is an observable of the 
domain playing the role of a hypothesis on the observation of an underlying 
process $\psi$. $\mathbf{A}_h$ represents a set of attributes, and $T^b_h$ and 
$T^e_h$ are two temporal instants representing the beginning and the end of the 
hypothesis. $m_1(\mathbf{A}_1,T_1), \ldots, m_n(\mathbf{A}_n,T_n)$ is a set of 
observables of the domain which plays the role of the evidence suggesting the 
observation of $h_\psi$. Each piece of evidence has its own set of attributes 
$\mathbf{A}_i$ and temporal support $T_i$, represented here as a single instant 
for the sake of simplicity, but it could also be an interval $(T^b_i, T^e_i)$. 
$C$ is a set of constraints among the variables involved in the abstraction 
pattern, which are interpreted as necessary conditions in order for the evidence 
$m_1(\mathbf{A}_1,T_1),\ldots,m_n(\mathbf{A}_n,T_n)$ to be abstracted into 
$h_\psi(\mathbf{A}_h,T^b_h, T^e_h)$. Finally, $\varTheta(\mathbf{A}_1, T_1, 
\ldots, \mathbf{A}_n, T_n)$ is an observation procedure that gives as a result 
an observation of $h_\psi(\mathbf{A}_h,T^b_h,T^e_h)$ from a set of observations 
for $m_1(\mathbf{A}_1,T_1),\ldots,$ $m_n(\mathbf{A}_n,T_n)$.

\begin{figure}[h]
 \centering
   \includegraphics[width=\textwidth]{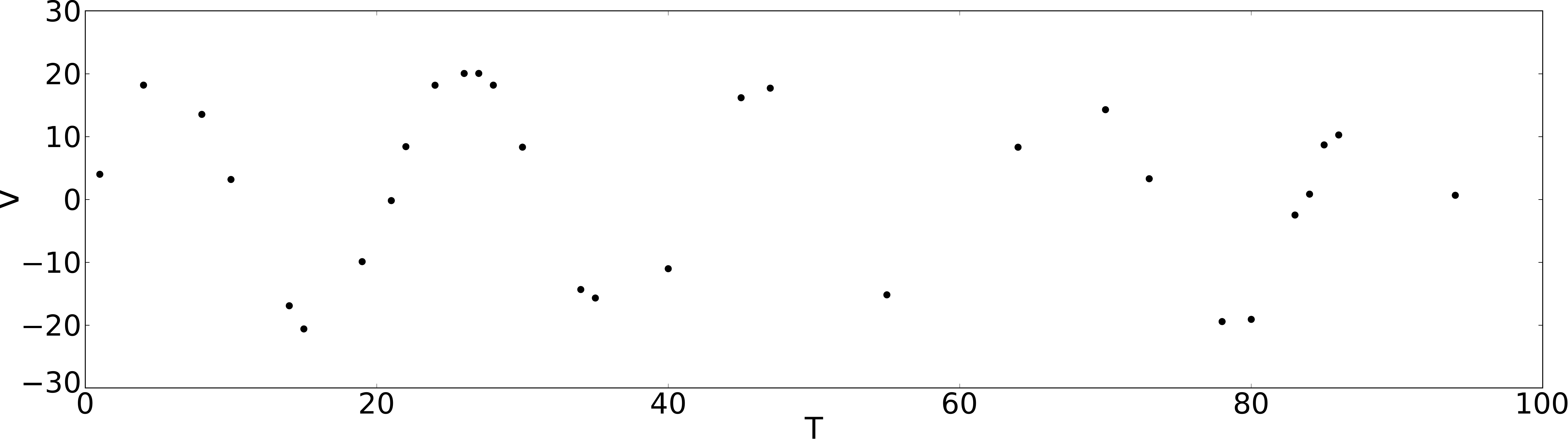}
   \resizebox{\textwidth}{!}{
\setlength{\tabcolsep}{3pt}
\renewcommand{\arraystretch}{1.2}
\begin{tabular}{c|rrrrrrrrrrrrrrrrrrrrrrrrrrrrrr}
$V$ & 3.4 & 17.6 & 12.9 & 2.6 & -17.5 & -20 & -10.5 & -0.8 & 7.8 
& 17.5 & 19.4 & 19.4 & 17.6 & 7.8 & -14.9 & -16.3 & -11.6 & 15.6 & 17.1 & -15.8 
& 7.7 & 13.7 & 2.7 & -19.8 & -19.6 & -3.1 & 0.2 & 8.1 & 9.6 & 0\\\hline
$T$ & 1 & 4 & 8 & 10 & 14 & 15 & 19 & 21 & 22 & 24 & 26 & 27 & 28 & 30 & 34& 35 
& 40 & 45 & 47 & 55 & 64 & 70 & 73 & 78 & 80 & 83 & 84 & 85 & 86 & 94
\end{tabular}}
  \caption{Initial temporal observations.}
  \label{fig:sinus_samples}  
\end{figure}

To illustrate this concept, consider the sequence of observations in 
Figure~\ref{fig:sinus_samples}. Each of these observations is an instance of an 
observable we call \textit{point} ($p$), represented as $p(\mathbf{A}=\{V\}, 
T)$, where $T$ determines the temporal location of the observation and $V$ is a 
value attribute.

If we analyze these observations visually, we may hypothesize the presence of an
underlying sinusoidal process. Let us define an observable $sinus$ for such a
sinusoidal process, with two attributes: the amplitude of the process ($\alpha$)
and its frequency ($\omega$). The knowledge necessary to conjecture this
hypothesis is collected in the following abstraction pattern:

\begin{eqnarray*}
[h_{sinus}(\{\alpha,\omega\}, T^b_h, T^e_h) = \varTheta(V_1,T_1,...,V_n,T_n)] 
& abstracts~ p(V_1,T_1),...,p(V_n,T_n)  \\
& \{C(\alpha,\omega,T^b_h,T^e_h,V_1,T_1,...,V_n,T_n)\}
\end{eqnarray*}

We can estimate the attribute values $(\alpha,\omega, T^b_h, T^e_h)$ of this 
process by a simple observation procedure $\varTheta$ that calculates $\alpha = 
max(|V_i|)$, for $1\leq i \leq n$, i.e., the amplitude $\alpha$ is obtained as 
the maximum absolute value of the observations; $\omega = 
\pi/mean(T^{peak}_j-T^{peak}_{j-1})$, where $T^{peak}_j$ are \textit{point} 
observations representing a peak, satisfying $(V^{peak}_j=V_k,T^{peak}_j=T_k) 
\wedge sign(V_{k}-V_{k-1})\neq sign(V_{k+1}-V_{k})$, so that the frequency 
$\omega$ is obtained as the inverse of the mean temporal separation between 
consecutive peaks in the sequence of observations; and $T^b_h = T_1,T^e_h=T_n$, 
i.e., the temporal support of the hypothesis is the time interval between the 
first and the last evidence points.

We can impose the following constraint 
$C(\alpha,\omega,T^b_h,T^e_h,V_1,T_1,\ldots,V_n,T_n)$ for every pair 
$(V_i,T_i)$ in the sequence:

\begin{displaymath}
 |\alpha \cdot sin(\omega \cdot T_i)-V_i| \leq \epsilon,
\end{displaymath}

\begin{figure}[h]  
 \centering
   \includegraphics[width=\textwidth]{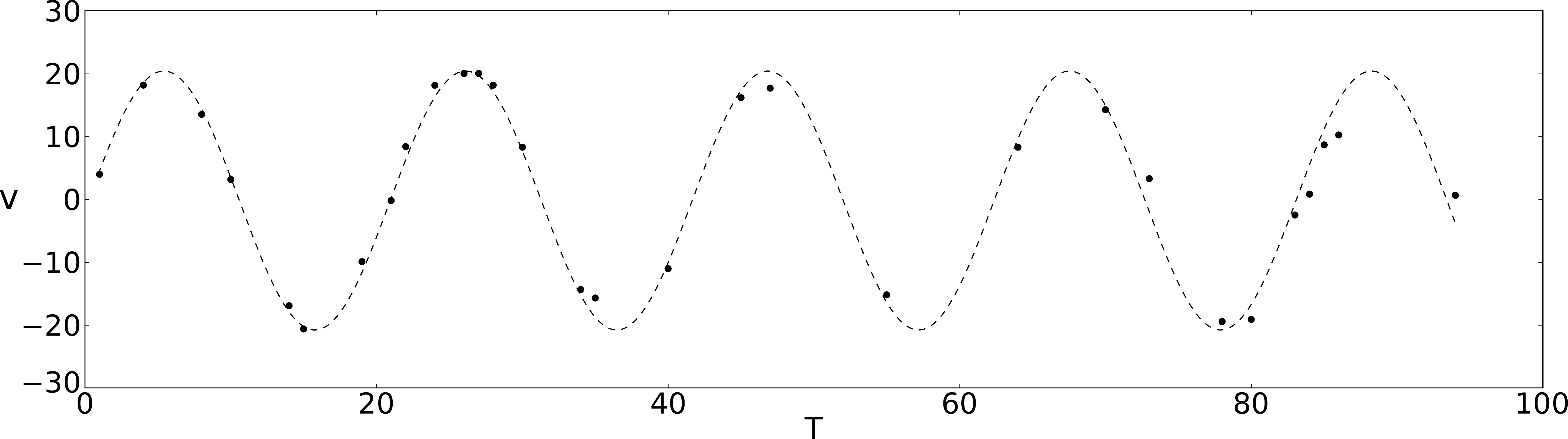}
  \caption{Abstracted sinusoidal process.}
  \label{fig:abstracted_sinus}  
\end{figure}

This constraint provides a model of a sinusoidal process and a measure of how 
well it fits a set of observations by means of a maximum error $\epsilon$. 
Figure~\ref{fig:abstracted_sinus} shows the continuous representation of the 
abstracted process, whose resulting observation is $h_{sinus}(\alpha = 20, 
\omega = 0.3, T^b_h = 1, T^e_h = 94)$. A value of $\alpha/3$ has been chosen 
for $\epsilon$. 

Of course, various observation procedures can be devised in order to estimate 
the same or different characteristics of the process being guessed. These 
procedures can provide one or several valid estimations in terms of their 
consistency with the abovementioned necessary constraints. In addition, 
different processes can be guessed from the same set of observations, all of 
them being valid in terms of their consistency. Hence, further criteria may be 
needed in order to rank the set of interpretations. 

This simple example summarizes the common approach to the interpretation of 
experimental results in science and technology, when the knowledge is available 
as a model or a set of models. The challenge is to assume that this knowledge is 
not available in an analytical but in a declarative form, as a pattern or a set 
of patterns, and that the interpretation task is expected to mimic certain 
mechanisms of human perception.

\section{Definitions}
\label{sec:definitions}
In this section we formally define the main pieces of our interpretation 
framework: {\em observables} and {\em observations} for representing the 
behavior of the system under study, and {\em abstraction patterns} for 
representing the knowledge about this system.

\subsection{Representation entities}
An {\em observation} is the result of measuring something with the quality of 
being {\em observable}. We call $\mathcal{Q}=\{q_0,q_1,...,q_n\}$ the set of
observables of a particular domain.

\begin{definition}
We define an \textbf{observable} as a tuple $q=\langle \psi,\mathbf{A},T^b, T^e 
\rangle$, where $\psi$ is a name representing the process being observable, 
$\mathbf{A}=\{A_1,...,A_{n_q}\}$ is a set of attributes to be valued, and $T^b$ 
and $T^e$ are two temporal variables representing the beginning and the end of 
the observable.
\end{definition}

We call $V_q(A_i)$ the domain of possible values for the attribute $A_i$. We 
assume a representation of the time domain $\tau$ isomorphic to the set of real 
numbers $\mathbb{R}$. In the case of an instantaneous observable, this is 
represented as $q=\langle \psi, \mathbf{A}, T \rangle$. Some observables can be 
dually represented from the temporal perspective, as either an observable 
supported by a temporal interval or as an observable supported by a temporal 
instant, according to the task to be carried out. A paradigmatic example is 
found in representing the heart beat, since it can be represented as a domain 
entity with a temporal extension comprising its constituent waves, and it can 
also be represented as an instantaneous entity for measuring heart rate.

\begin{example}
{\em In the ECG signal, several distinctive waveforms can be identified, 
corresponding to the electrical activation-recovery cycle of the different heart 
chambers. The so-called P wave represents the activation of the atria, and is 
the first wave of the cardiac cycle. The next group of waves recorded is the QRS 
complex, representing the simultaneous activation of the right and left 
ventricles. Finally, the wave that represents the ventricular recovery is called 
the T wave. Together, these waveforms devise the characteristic pattern of the 
heart cycle, which is repeated in a normal situation with every 
beat~\cite{Marriott08}. An example of a common ECG strip is shown in 
Figure~\ref{fig:ecg}.

According to this description, the observable $q_{Pw}=\langle 
\mathtt{atrial\_activation},$ $\{\mathtt{amplitude}\}, T^b, T^e \rangle$ 
represents a P wave resulting from an atrial activation process with an unknown 
amplitude, localized in a still unknown temporal interval. 
}

\begin{figure}[ht!]  
 \centering
   \includegraphics[width=\textwidth]{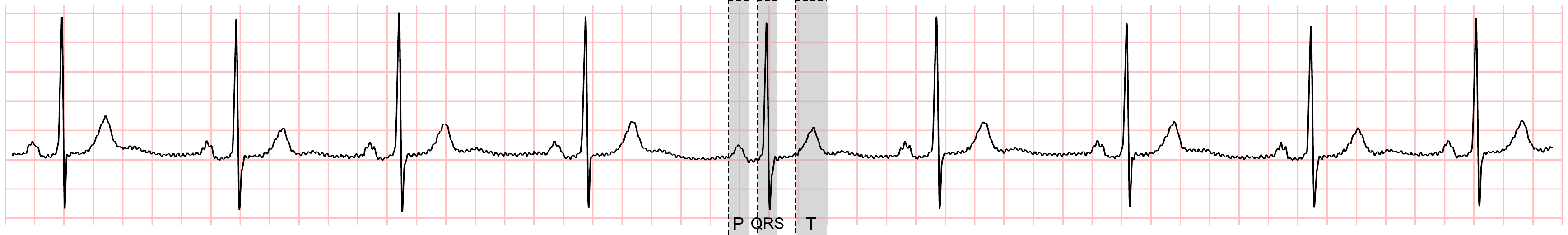}
  \caption{Example of the ECG basic waveforms. [Source: MIT-BIH arrhythmia 
DB~\cite{Goldberger00}, recording: 123, between 12:11.900 and 12:22.400]}
  \label{fig:ecg}  
\end{figure}

\end{example}

\begin{definition}
 We define an \textbf{observation} as a tuple $o=\langle q, \mathbf{v}, t^b, 
t^e\rangle$, an instance of the observable $q$ resulting from assigning a 
specific value to each attribute and to the temporal variables, where 
$\mathbf{v}=(v_1, \ldots,v_{n_q})$ is the set of attribute values such that 
$\mathbf{v}\in V_q(A_1)\times \ldots \times V_q(A_{n_q})$ and $t^b,~t^e\in \tau$ 
are two precise instants limiting the beginning and the end of the observation.
\end{definition}
We also use the notation $(A_1=v_1, \ldots, A_{n_q}=v_{n_q})$ to represent 
the assignment of values to the attributes of the observable and $T^b=t^b$ and 
$T^e=t^e$ for representing the assignment of temporal limits to the observation.

\begin{example}
{\em The tuple $o=\langle q_{Pw},\mathtt{0.17mV}, 
\mathtt{12:16.977}, \mathtt{12:17.094}\rangle$ represents the particular 
occurrence of the P wave observable highlighted in Figure~\ref{fig:ecg}.}
\end{example}

Some notions involving observables and observations are defined below that will 
be useful in describing certain properties and constraints of the domain 
concepts, as well as in temporally arranging the interpretation process.

\begin{definition}
 Given a set of observables $\mathcal{Q}$, a \textbf{generalization relation} 
can be defined between two different observables $q=\langle \psi,\mathbf{A},T^b, 
T^e \rangle$ and $q'=\langle \psi',\mathbf{A}',$ $T'^b, T'^e \rangle$, denoted 
by $q' ~ \mathtt{is ~a} ~ q$, meaning that $q$ generalizes $q'$ if and only if  
$\mathbf{A} \subseteq \mathbf{A'}$ and $V_{q'}(A_i)\subseteq V_q(A_i)~ \forall 
A_i\in \mathbf{A}$.
\end{definition}

The generalization relation is reflexive, antisymmetric and transitive. The 
inverse of a generalization relation is a specification relation. From a logical 
perspective, a generalization relation can be read as an implication $q' 
\rightarrow q$, meaning that $q'$ is more specific than $q$. It holds that 
every observation $o=\langle q',\mathbf{v},t^b,t^e\rangle$ of the observable 
$q'$ is also an observation of $q$.

\begin{example}
 {\em A common example of a generalization relation can be defined from a domain 
partition of an attribute. For example, $q_1 = \langle \mathtt{Sinus\_Rhythm},$ 
$\{\mathtt{RR} \in [200ms, 4000ms]\}, T^b, T^e\rangle$ is a generalization of 
the observables $q_2 = \langle \mathtt{Sinus\_Tachycardia}, \{\mathtt{RR} \in 
[200ms, 600ms]\}, T^b, T^e\rangle$, $q_3 =\langle \mathtt{Normal\_Rhythm},$ 
$\{\mathtt{RR} \in [600ms, 1000ms]\}, T^b, T^e\rangle$ and $q_4 = \langle 
\mathtt{Sinus\_Bradycardia},$ $\{\mathtt{RR} \in [1000ms,$ $ 4000ms]\}, T^b, 
T^e\rangle$. The \texttt{RR} attribute represents the measure of the mean time 
distance between consecutive beats, while $q_2, q_3$ and $q_4$ represent the 
normal cardiac rhythm denominations according to the heart 
rate~\cite{Marriott08}.}
\end{example}

\begin{definition} 
\label{def:exclusion}
Given a set of observables $\mathcal{Q}$, an \textbf{exclusion relation} can be 
defined between two different observables $q=\langle \psi,\mathbf{A},T^b, T^e 
\rangle$ and $q'=\langle \psi',\mathbf{A}',$ $T'^b, T'^e \rangle$, denoted by $q 
~ \mathtt{excludes} ~ q'$, meaning that they are mutually exclusive  if and only 
if their respective processes $\psi$ and $\psi'$ cannot concurrently occur. 
\end{definition}

The exclusion relation is defined by extension from the knowledge of the domain, 
and its rationale lies in the nature of the underlying processes and mechanisms. 
Inasmuch as the occurrence of a process can only be hypothesized as long as it 
is observable, the exclusion relation behaves as a restriction on observations. 
Thus, given two observables $q$ and $q'$, $q ~ \mathtt{excludes} ~ q'$ entails 
that they cannot be observed over two overlapping intervals, i.e., every two 
observations $o=\langle q,\mathbf{v},t^b,t^e\rangle$ and $o'=\langle 
q',\mathbf{v}',t'^b,t'^e\rangle$ satisfy either $t^e < t'^b$ or $t'^e < t^b$. 
The opposite is not generally true. The exclusion relation is symmetric and 
transitive. As an example, in the domain of electrocardiography, the knowledge 
about the physiology of the heart precludes the observation of a \textit{P wave} 
during an episode of \textit{Atrial fibrillation}~\cite{Marriott08}, so these 
two observables are mutually exclusive.

We call $\mathcal{O}$ the set of observations available for the observables in 
$\mathcal{Q}$. In order to index this set of observations, they will be 
represented as a sequence by defining an order relation between them. This 
ordering aims to prioritize the interpretation of the observations as they 
appear. 

\begin{definition}
 Let $<$ be an \textbf{order relation} between two observations $o_i=\langle
q_i,\mathbf{v}_i,t^b_i,t^e_i\rangle$ and $o_j=\langle
q_j,\mathbf{v}_j,t^b_j,t^e_j\rangle$ such that $(o_i<o_j) \Leftrightarrow (t_i^b
< t_j^b) \vee ((t_i^b = t_j^b)\wedge (t_i^e < t_j^e)) \vee ((t_i^b =
t_j^b)\wedge (t_i^e = t_j^e)\wedge (q_i<q_j))$, assuming a lexicographical order
between observable names. 
\end{definition}

A sequence of observations is an ordered set of observations 
$\mathcal{O}=(o_1,...,o_i,...)$ where for all $i<j$ then $o_i< o_j$. Every 
subset of a sequence of observations is also a sequence. The q-sequence of 
observations from $\mathcal{O}$, denoted as $O(q)$, is the subset of the 
observations for the observable $q$. The exclusion relation forces that any two 
observations $o_i=\langle q,\mathbf{v}_i,t^b_i,t^e_i\rangle$ and $o_j=\langle 
q,\mathbf{v}_j,t^b_j,t^e_j\rangle$ in $O(q)$ satisfy $o_i < o_j \Rightarrow 
t^e_i< t^b_j$ for the current application domain. By $\mathtt{succ}(o_i)$ we 
denote the successor of the observation $o_i$ in the sequence $\mathcal{O}$, 
according to the order relation $<$. By q-$\mathtt{succ}(o_i)$ we denote the 
successor of the observation $o_i\in O(q)$ in its q-sequence $O(q)$. Conversely 
to this notation, we denote by $q(o_i)$ the observable corresponding to the 
$o_i$ observation.

\subsection{Abstraction patterns}

We model an abstraction process as an abduction process, based on the 
conjectural relation $m ~\leftarrow~ h$ \cite{Peirce31}, which can be read as 
{\em `the observation of the finding $m$ allows us to conjecture the observation 
of $h$ as a possible explanatory hypothesis'}. For example, a very prominent 
peak in the ECG signal allows us to conjecture the observation of a heartbeat. A 
key aspect of the present proposal is that both the hypothesis and the finding 
are observables, and therefore formally identical, i.e., there exists 
$q_i,q_j\in \mathcal{Q}$, with $q_i\neq q_j$, such that $h\equiv q_i=\langle 
\psi_i,\mathbf{A}_i,T_i^b, T_i^e \rangle$ and $m\equiv q_j=\langle 
\psi_j,\mathbf{A}_j,T_j^b, T_j^e \rangle$. In general, an abstraction process 
can involve a number of different findings, even multiple findings of the same 
observable, and a set of constraints among them; thus, for example, a regular 
sequence of normal heartbeats allows us to conjecture the observation of a sinus 
rhythm. Additionally, an observation procedure is required in order to produce 
an observation of the hypothesis from the observation of those findings involved 
in the abstraction process.

We devise an abstraction process as a knowledge-based reasoning process, 
supported by the notion of abstraction pattern, which brings together those 
elements required to perform an abstraction. Formally:
\begin{definition}
An \textbf{abstraction pattern} $P=\langle h,M_P,C_P,\varTheta_P\rangle$ 
consists of a hypothesis $h$, a set of findings $M_P = \{m_1,\ldots,m_n\}$, a 
set of constraints $C_P=\{C_1, \ldots,C_t\}$ among the findings and the 
hypothesis, and an observation procedure $\varTheta_P(\mathbf{A}_1, T^b_1, 
T^e_1, \ldots,\mathbf{A}_n, T^b_n, T^e_n)\in O(h)$.
\end{definition}

Every constraint $C_i\in C_P$ is a relation defined on a subset of the set of 
variables taking part in the set of findings and the hypothesis $
\{\mathbf{A}_h, T^b_h, T^e_h, \mathbf{A}_1, $ $T^b_1, T^e_1, \ldots, 
\mathbf{A}_n, T^b_n, T^e_n\}$. Thus, a constraint is a subset of the Cartesian 
product of the respective domains, and represents the simultaneously valid 
assignments to the variables involved. We will denote each constraint by 
making reference to the set of variables being constrained, as in 
$C_P(\mathbf{A}_h,T^b_h,T^e_h,\mathbf{A}_1,$ $T^b_1,T^e_1,\ldots, \mathbf{A}_n, 
T^b_n, T^e_n)$ for the whole abstraction pattern. 

An abstraction pattern establishes, through the set $C_P$, the conditions for 
conjecturing the observation of $h$ from a set of findings $M_P$, and through 
the observation procedure $\varTheta_P$, the calculations for producing a new 
observation $o_h \in O(h)$ from the observation of these findings. We call 
$M^q_P=\{m^q_1,m^q_2,...,m^q_s\}$ the set of findings of the observable $q$ in 
$P$, being $M_P=\bigcup _{q\in \mathcal{Q}} M^q_P$. Thus, a set of findings 
allows the elements of a multiset of observables to be distinguished. The 
interpretation procedure will choose, as we will see later, from the available 
observations for every observable $q$ satisfying the constraints $C_P$, which 
are to be assigned to the findings in $M^q_P$ in order to calculate $o_h$.

The set of findings $M_P$ is divided into two disjoint sets $A_P$ and $E_P$, 
where $A_P$ is the set of findings that is said to be abstracted in $o_h$, and 
$E_P$ is the set of findings that constitute the observation environment of 
$o_h$, that is, the set of findings needed to properly conjecture $o_h$, but 
which are not synthesized in $o_h$.

A temporal covering assumption can be made as a \textit{default 
assumption}~\cite{Poole90} on a hypothesis $h=\langle \psi_h, \mathbf{A}_h, 
T_h^b, T_h^e \rangle$ with respect to those findings $m=\langle \psi_m, 
\mathbf{A}_m,$ $T_m^b, T_m^e \rangle$ appearing in an abstraction pattern:

\begin{defassumption}
 \emph(Temporal covering)  Given an abstraction pattern $P$, it holds that
$T_h^b \leq T_m^b$ and $T_m^e \leq T_h^e$, for all $m\in A_P\subseteq M_P$.
\end{defassumption}

The temporal covering assumption allows us to define the exclusiveness of an
interpretation as the impossibility of including competing abstractions in the
same interpretation.

\begin{example}
{\em \label{ex:wave} According to~\cite{CSE85}, in the electrocardiography 
domain a \textit{``wave''} is a discernible deviation from a horizontal 
reference line called baseline, where at least two opposite slopes can be 
identified. The term discernible means that both the amplitude and the duration 
of the deviation must exceed some minimum values, agreed as 20 
\textmu V and 6 ms respectively. A wave can be completely described by a set of 
attributes: its amplitude ($A$), voltage polarity ($VP \in \{+,-\}$) and its 
main turning point $T^{tp}$, resulting in the following observable: }
\begin{displaymath}
q_{wave}=\langle\mathtt{electrical\_activity}, \{A,VP,T^{tp}\}, T^b, T^e\rangle
\end{displaymath}

{\em Let us consider the following abstraction pattern:}

\begin{align*}
 P_{wave} &= \langle wave, M_{P} = \{m^{ECG}_0,\ldots,m^{ECG}_n\}, C_{P_{wave}}, 
\mathtt{wave\_observation()} \rangle
\end{align*}
{\em where $m^{ECG}_i$ is a finding representing an ECG sample, with a single 
attribute $V_i$ representing the sample value, and a temporal variable $T_i$ 
representing its time point. We set the onset and end of a wave to the time of 
the second $m^{ECG}_1$ and second-to-last $m^{ECG}_{n-1}$ samples, considering  
$m^{ECG}_0$ and $m^{ECG}_n$ as environmental observations which are used to 
check the presence of a slope change just before and after the wave; thus 
$E_{P_{wave}} = \{m^{ECG}_0, m^{ECG}_n\}$, and $A_{P_{wave}} = \{m^{ECG}_1, 
\ldots, m^{ECG}_{n-1}\}$.}

{\em A set of temporal constraints are established between the temporal 
variables: $c_{1}=  \{ T^e - T^b \geq 6ms \}$, $c_2= \{T^b= T_1\}$, $c_3= \{T^e 
= T_{n-1}\}$ and $c_4= \{T^b < T^{tp} < T^e\}$. Another set of constraints 
limit the amplitude and slope changes of the samples included in a wave: $c_5= 
\{ sign(V_1 - V_0) \neq sign(V_2 - V_1)\}$, $c_6= \{ sign(V_{n} - V_{n-1}) \neq 
sign(V_{n-1} - V_{n-2})\}$, $c_7= \{sign(V_{tp} - V_{tp-1}) = -sign(V_{tp+1} - 
V_{tp})\}$ and $c_{8}= \{min\{|V_{tp}-V_1|, |V_{tp}-V_{n-1}|\} \geq 20 \mu V\}$. 
These two sets form the complete set of constraints of the pattern 
$C_{P_{wave}} = \{c_1,\ldots,c_8\}$.
}

{\em Once a set of ECG samples has satisfied these constraints, they support the
observation of a wave: $o_{wave}=\langle q_{wave}, (a, vp, t^{tp}), t^b, 
t^e\rangle$. The values of $t^b$ and $t^e$ are completely determined by the 
constraints $c_2$ and $c_3$, while the observation procedure 
$\mathtt{wave\_observation()}$ provides a value for the attributes as follows: 
$vp= sign(V_{tp}-V_1)$, $a= max\{|V_{tp}-V_1|,|V_{tp}-V_{n-1}|\}$, and $t^{tp} = 
t^b + tp$, where $tp = arg\,min _k \{V_k|1\leq k\leq {n-1}\}$, if $V_1 < V_0$, 
or $tp = arg\,max_k \{V_k|1\leq k\leq {n-1}\}$, if $V_1 > V_0$.} 

\end{example}

\subsection{Abstraction grammars}

According to the definition, an abstraction pattern is defined over a fixed set 
of evidence findings $M_P$. In general, however, an abstraction involves an 
undetermined number of pieces of evidence (in the case of an ECG wave, the 
number of samples). Hence, we provide a procedure for dynamically generating 
abstraction patterns, based on the theory of formal languages. The set 
$\mathcal{Q}$ of observables can be considered as an alphabet. Given an alphabet 
$\mathcal{Q}$, the special symbols $\varnothing$ (empty set), and $\lambda$ 
(empty string), and the operators $|$ (union), $\cdot$ (concatenation), and 
$\ast$ (Kleene closure), a formal grammar $G$ denotes a pattern of symbols of 
the alphabet, describing a language $L(G)\subseteq \mathcal{Q}^\ast$ as a subset 
of the set of possible strings of symbols of the alphabet.

Let $G^{ap}$ be the class of formal grammars of abstraction patterns. An {\em 
abstraction grammar} $G \in G^{ap}$ is syntactically defined as a tuple $(V_N, 
V_T, H, R)$. For the production rules in $R$ the expressiveness of right-linear 
grammars is adopted \cite{Hopcroft01}:
\begin{align}
\label{eq:pat_grammar}
H &\rightarrow q D \nonumber\\
D &\rightarrow q F~|~q ~|~\lambda \nonumber
\end{align}

$H$ is the initial symbol of the grammar, and this plays the role of the 
hypothesis guessed by the patterns generated by $G$. $V_N$ is the set of 
non-terminal symbols of the grammar, satisfying $H\in V_N$, although $H$ cannot 
be found on the right-hand side of any production rule, since a hypothesis 
cannot be abstracted by itself. $V_T$ is the set of terminal symbols of the 
grammar, representing the set of observables $Q_G\subseteq \mathcal{Q}$ that can 
be abstracted by the hypothesis.

Given a grammar $G \in G^{ap}$, we devise a constructive method for generating a 
set of abstraction patterns $P_G=\{P_1,\ldots,P_i,\ldots\}$. Since a formal 
grammar is simply a syntactic specification of a set of strings, every grammar 
$G \in G^{ap}$ is semantically extended to an attribute grammar \cite{Aho06}, 
embedded with a set of actions to be performed in order to incrementally build 
an abstraction pattern by the application of production rules. An abstraction 
grammar is represented as $G=((V_N,V_T,H,R),B, BR)$, where $B(\alpha)$ 
associates each grammar symbol $\alpha \in V_N\cup V_T$ with a set of 
attributes, and $BR(r)$ associates each rule $r\in R$ with a set of attribute 
computation rules. An abstraction grammar associates the following attributes: 
{\em i)} $P(attern)$, with each non-terminal symbol of the grammar; this will be 
assigned an abstraction pattern; {\em ii)} $A(bstracted)$, with each terminal 
symbol corresponding to an observable $q\in Q_G$; this allows us to assign each 
finding either to the set $A_P$ or $E_P$, depending on its value of true or 
false; {\em iii)} $C(onstraint)$, with each terminal symbol corresponding to an 
observable; this will be assigned a set of constraints. There are approaches in 
the bibliography dealing with different descriptions of Constraint Satisfaction 
Problems and their semantic expression in different formalisms 
\cite{Barro94a,Chakravarty00,Dechter03}. By explicitly specifying a constraint 
as a relation a clear description is provided on its underlying meaning, but 
this can lead to cumbersome knowledge representation processes. Multiple 
mathematical conventions can concisely and conveniently describe a constraint as 
a Boolean-valued function over the variables of a set of observables. However, 
we will focus on the result of applying a set of constraints among the variables 
involved.

In the following, the set of attribute computation rules associated with the 
grammar productions is specified to provide a formal method for building 
abstraction patterns $P\in P_{G_h}$ from a grammar $G_h \in G^{ap}$. $P_{G_h}$ 
gathers the set of abstraction patterns that share the same observable $h$ as a 
hypothesis; thus, these represent the different ways to conjecture $h$. Using 
this method, the application of every production incrementally adds a new 
observable as a finding and a set of constraints between this finding and 
previous entities, as follows: 

\begin{enumerate}
\item  The initial production $H \rightarrow q D$ entails:
\begin{eqnarray}
P_H & := & \langle h, M_H=\varnothing, C_H = \varnothing, \varTheta_H= 
\varnothing \rangle \nonumber \\
C_q & := & C(\mathbf{A}_h,T^b_h, T^e_h,\mathbf{A}_1,T^b_1, T^e_1) \nonumber \\
A_q & \in & \{true,false\} \nonumber \\
P_D & := &  \langle h, M_D=  M_H\cup 
\{m_1^q\}, C_D=C_H \cup C_q, \varTheta_D(\mathbf{A}_1,T^b_1,T^e_1) \rangle 
\nonumber
\end{eqnarray}
\item All productions in the form $D \rightarrow q F$  entail:
\begin{eqnarray}
P_D & := & \langle h, M_D, C_D, \varTheta_D(\mathbf{A}_1, T^b_1, T^e_1, \ldots, 
\mathbf{A}_{k}, T^b_{k}, T^e_{k}) \rangle \nonumber \\
C_q & := & C(\mathbf{A}_h,T^b_h, T^e_h, \mathbf{A}_1, \ldots, 
\mathbf{A}_{k+1},T^b_{k+1}, T^e_{k+1}) \nonumber \\
A_q & \in & \{true,false\} \nonumber \\
P_F & := & \langle h, M_F=M_D \cup \{m^q_{k+1}\},C_F= C_D \cup C_q, 
\varTheta_F(\mathbf{A}_1,T^b_1,T^e_1,\ldots,\mathbf{A}_{k+1},T^b_{k+1},T^e_{k+1}
) \rangle \nonumber
\end{eqnarray}
\item Productions in the form $D \rightarrow q$ conclude the generation of a 
pattern $P\in P_{G_h}$:
\begin{eqnarray}
P_D & := & \langle h, M_D, C_D, \varTheta_D(\mathbf{A}_1, T^b_1, T^e_1, \ldots, 
\mathbf{A}_{k}, T^b_{k}, T^e_{k}) \rangle \nonumber \\
C_q & := & C(\mathbf{A}_h,T^b_h, T^e_h, \mathbf{A}_1, \ldots, \mathbf{A}_{k+1}, 
T^b_{k+1}, T^e_{k+1}) \nonumber \\
A_q & \in & \{true,false\} \nonumber \\
P & := & \langle h, M_P=M_D \cup \{m^q_{k+1}\},C_P=C_D \cup C_q, 
\varTheta_P(\mathbf{A}_1, T^b_1, T^e_1, \ldots, \mathbf{A}_{k+1}, T^b_{k+1}, 
T^e_{k+1}) \rangle  \nonumber
\end{eqnarray}
\item Productions in the form $D \rightarrow \lambda$ also conclude the 
generation of a pattern:
\begin{eqnarray}
P_D & := & \langle h, M_D, C_D, \varTheta_D(\mathbf{A}_1,T^b_1,T^e_1, \ldots, 
\mathbf{A}_k,T^b_k,T^e_k) \rangle \nonumber \\
P & := & P_D \nonumber
\end{eqnarray}
\end{enumerate}

This constructive method enables the incremental addition of new constraints as 
new findings are included in the representation of the abstraction pattern, 
providing a dynamic mechanism for knowledge assembly by language generation. The 
final constraints in $C_P$ are obtained from the conjunction of the constraints 
added at each step. Moreover, it is possible to design an adaptive observation 
procedure as new evidence becomes available, since the observation procedure may 
be different at each step.

In the case that no temporal constraints are attributed to a production, a 
'hereafter' temporal relationship will be assumed by default to exist between 
the new finding and the set of previous findings. For instance, a production of 
the form $D \rightarrow qF$  entails that $C_F= C_P \cup \{T^b_i \leq T^b_{k+1} 
~|~ m_i \in M_P \}$.

Hence, in the absence of any temporal constraint, an increasing temporal order 
among consecutive findings in every abstraction pattern is assumed. Moreover, 
every temporal constraint must be consistent with this temporal order.

According to the limitation imposed on observations of the same observable which 
prevents two different observations from occurring at the same time, an 
additional constraint is added on any two findings of the same observable, and 
thus $\forall m^q_i,m^q_j \in M^q_P, (T^e_i < T^{b}_j \vee T^e_j <
T^b_i)$.

Several examples of abstraction pattern grammars modeling common knowledge in 
electrocardiography are given below, in order to illustrate the expressiveness 
of the $G^{ap}$ grammars.

\begin{example}
\label{ex:ncy}
{\em The grammar $G_{N}=(V_N,V_T,H,R)$ is designed to generate an abstraction 
pattern for a \textit{normal cardiac cycle}, represented by the observable 
$q_N$, including the descriptions of common durations and 
intervals~\cite{Marriott08}. In this grammar, $V_N=\{H,D,E\}$, $V_T=$ 
$\{q_{Pw}, q_{QRS}, q_{Tw}\}$, and $R$ is given by:}

\begin{align*}
H &\rightarrow q_{Pw} D & & {\scriptstyle \{P_H :=~ \langle q_N, 
 M_H=\varnothing, C_H = \varnothing,\varTheta_H= \varnothing \rangle,}\\
 & & &~ {\scriptstyle C_{Pw} :=~ \{T^b_N = T^b_{Pw};~ 50ms \leq T^e_{Pw} 
-  T^b_{Pw} \leq 120ms\},}\\
 & & &~ {\scriptstyle A_{Pw} :=~ true,}\\
 & & &~ {\scriptstyle P_D :=~ \langle q_N, M_D=\{m^{Pw}\}, C_D=C_{Pw}, 
\varTheta_D=\varnothing \rangle}\\[-5pt]
 & & & \scriptstyle\}\\
D &\rightarrow q_{QRS} E & & {\scriptstyle \{P_D :=~ \langle q_N, 
 M_D=\{m^{Pw}\}, C_D=C_{Pw}, \varTheta_D=\varnothing \rangle,}\\ 
 & & &~ {\scriptstyle C_{QRS} :=~ \{50ms \leq T^e_{QRS} - T^b_{QRS} \leq 
 150ms;~ 100ms \leq T^b_{QRS} - T^b_{Pw} \leq 210ms \},}\\
 & & &~ {\scriptstyle A_{QRS} :=~ true,}\\
 & & &~ {\scriptstyle P_E :=~ \langle q_N, M_E=M_D\cup\{m^{QRS}\}, C_E= C_D 
 \cup C_{QRS}, \varTheta_E=\varnothing \rangle}\\[-5pt]
 & & & \scriptstyle\}\\ 
E &\rightarrow q_{Tw} & & {\scriptstyle \{P_E :=~ \langle q_N, 
 M_E=\{m^{Pw}, m^{QRS}\}, C_E, \varTheta_E =\varnothing \rangle,}\\
 & & &~ {\scriptstyle C_{Tw} :=~ \{80ms \leq T^b_{Tw} - T^e_{QRS} \leq 
120ms;~ T^e_{Tw} - T^b_{QRS} \leq 520ms;~ T^e_N = T^e_{Tw}\},}\\
 & & &~ {\scriptstyle A_{Tw} :=~ true,}\\
 & & &~ {\scriptstyle P :=~ \langle q_N, M_P=M_E\cup\{m^{Tw}\}, C_P= C_E \cup 
 C_{Tw}, \varTheta_P = \varnothing \rangle}\\[-5pt]
 & & & \scriptstyle\}
\end{align*}
{\em 
This grammar generates a single abstraction pattern, which allows us to 
interpret the sequence of a P wave, a QRS complex, and a T wave as the 
coordinated contraction and relaxation of the heart muscle, from the atria to 
the ventricles. Some additional temporal constraints are required and specified 
in the semantic description of the production rules. In this case, an 
observation procedure $\varTheta$ is not necessary since the attributes of the 
hypothesis are completely determined by the constraints in the grammar, and do 
not require additional calculus.}
\end{example}

The next example shows the ability of an abstraction grammar to 
generate abstraction patterns dynamically with an undefined number of findings.

\begin{example}
\label{ex:vbigeminy}
{\em A \textit{bigeminy} is a heart arrhythmia in which there is a continuous 
alternation of long and short heart beats. Most often this is due to ectopic 
heart beats occurring so frequently that there is one after each normal beat, 
typically premature ventricular contractions (PVCs)~\cite{Marriott08}. For 
example, a normal beat is followed shortly by a PVC, which is then followed by a 
pause. The normal beat then returns, only to be followed by another PVC. The 
grammar $G_{VB} = (V_N, V_T, H, R)$ generates a set of abstraction patterns for 
\textit{ventricular bigeminy}, where $V_N=\{H,D,E,F\}$, $V_T=$ $\{q_N,q_V\}$, 
and $R$ is given by:}

\begin{align*}
H &\rightarrow q_N D & & {\scriptstyle \{ P_H :=~ \langle q_{VB}, 
 M_H=\varnothing, C_H = \varnothing,\varTheta_H= \varnothing \rangle,}\\ 
 & & &~ {\scriptstyle C_N :=~ \{ T^b_{VB} = T_1 \},}\\
 & & &~ {\scriptstyle A_N :=~ true,}\\
 & & &~ {\scriptstyle P_D :=~ \langle q_{VB}, M_D=\{m^{N}_1\}, C_D=C_N, 
\varTheta_D = \varnothing \rangle}\\[-5pt]
 & & & \scriptstyle\}\\
D &\rightarrow q_V E & & {\scriptstyle \{ P_D :=~ \langle q_{VB}, 
 M_D=\{m^{N}_1\}, C_D=C_N, \varTheta_D = \varnothing\rangle,}\\
 & & &~ {\scriptstyle C_{V} :=~ \{ 200ms \leq T_2 - T_1 \leq 800ms \},}\\
 & & &~ {\scriptstyle A_{V} :=~ true,}\\
 & & &~ {\scriptstyle P_E :=~ \langle q_{VB}, M_E=M_D \cup \{m^{V}_2\}, 
 C_E=C_D \cup C_{V}, \varTheta_E = \varnothing \rangle}\\[-5pt]
 & & & \scriptstyle\}\\
E &\rightarrow q_N F & & {\scriptstyle \{ P_E :=~ \langle q_{VB}, 
 M_E=\{m^{N}_1, \ldots, m^{V}_{k-1}\}, C_E,\varTheta_E=\varnothing \rangle,}\\ 
 & & &~ {\scriptstyle C_{N} :=~ \{ 1.5 \cdot 200ms \leq T_k - T_{k-1} 
 \leq 4 \cdot 800ms \},}\\
 & & &~ {\scriptstyle A_{N} :=~ true,}\\
 & & &~ {\scriptstyle P_F :=~ \langle q_{VB}, M_F=M_E\cup\{m^{N}_k\}, C_F=C_E 
 \cup C_{N}, \varTheta_F = \varnothing \rangle}\\[-5pt]
 & & & \scriptstyle\}\\
F &\rightarrow q_V E & & {\scriptstyle \{ P_F :=~ \langle q_{VB}, M_F=\{m^{N}_1, 
 m^V_2, \ldots, m^{N}_{k}\}, C_F,\varTheta_F = \varnothing \rangle,}\\ 
 & & &~ {\scriptstyle C_{V} :=~ \{200ms \leq T_{k+1} - T_k \leq 800ms\},}\\
 & & &~ {\scriptstyle A_{V} :=~ true,}\\
 & & &~ {\scriptstyle P_E :=~ \langle q_{VB}, M_E=M_F\cup\{m^{V}_{k+1}\}, 
 C_E=C_F \cup C_{V}, \varTheta_F = \varnothing \rangle}\\[-5pt]
 & & & \scriptstyle\}\\
F &\rightarrow q_V & & {\scriptstyle \{ P_F :=~ \langle q_{VB}, M_F=\{m^{N}_1, 
 m^V_2, \ldots, m^{N}_{n-1}\}, C_F,\varTheta_F = \varnothing \rangle,}\\ 
 & & &~ {\scriptstyle C_{V} :=~ \{ 200ms \leq T_n - T_{n-1} \leq 800ms;~ 
T^e_{VB} = T_n \},}\\
 & & &~ {\scriptstyle A_{V} :=~ true,}\\
 & & &~ {\scriptstyle P :=~ \langle q_{VB}, M_P=M_F\cup\{m^{V}_n\}, C_P=C_F 
\cup C_{V}, \varTheta_P = \varnothing \rangle}\\[-5pt]
 & & & \scriptstyle\}
\end{align*}

{\em
For simplicity, we have referenced each $N$ and $V$ heart beat with a single 
temporal variable. Thus $T_i$ represents the time point of the ith heart beat, 
and is a normal beat if $i$ is odd, and a PVC if $i$ is even. With the execution 
of these production rules, an unbounded sequence of alternating normal and 
premature ventricular QRS complexes is generated, described above as 
ventricular bigeminy. Note that in terms of the $\{N, V\}$ symbols the $G_{VB}$ 
grammar is syntactically equivalent to the regular expression $NV(NV)^+$. 

In this example, as in~\ref{ex:ncy}, an observation procedure $\varTheta_P$ is 
not necessary, since the constraints in the grammar completely determine the 
temporal endpoints of the hypothesis and there are no more attributes to be 
valued. Figure~\ref{fig:vb_figure} shows an example of a ventricular bigeminy 
pattern.
}

\begin{figure}[h!]  
 \centering
  \includegraphics[width=\textwidth]{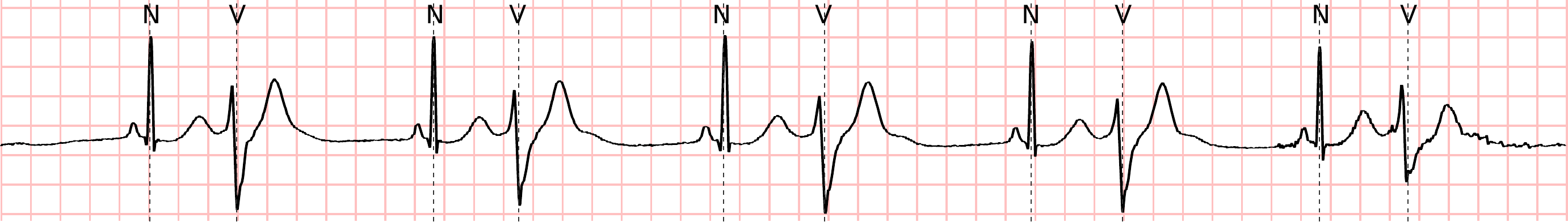}
 \caption{Example of ventricular bigeminy. [Source: MIT-BIH arrhythmia DB, 
recording: 106, between 25:06.350 and 25:16.850]}
 \label{fig:vb_figure}  
\end{figure}
\end{example}

\section{An interpretation framework}
\label{sec:framework}
In this section, we define and characterize an interpretation problem. 
Informally, an interpretation problem arises from the availability of a set of 
initial observations from a given system, and of domain knowledge formalized as 
a set $\mathcal{G}=\{G_{q_1},\ldots,G_{q_n}\}$ of $G^{ap}$ grammars. Every 
abstraction grammar $G_h \in \mathcal{G}$ generates a set of abstraction 
patterns that share the same hypothesis $h$. The whole set of abstraction 
patterns that can be generated by $\mathcal{G}$ is denoted as $\mathcal{P}$.

\begin{definition}
 Let $\mathcal{Q}$ be a set of observables and $\mathcal{G}$ a set of
abstraction grammars. We say $\mathcal{G}$ induces an \textbf{abstraction
relation} in $\mathcal{Q} \times \mathcal{Q}$, denoted by $q_i~\crel~q_j$ if
and only if there exists an abstraction pattern $P$ generated by some $G_h\in
\mathcal{G}$ such that:
\begin{enumerate}
 \item $q_j = h$
 \item $M^{q_i}_P \cap A_P \neq \varnothing$
 \item $q_i \ncrel^+ q_i$, where $\crel^+$ is the transitive closure of $\crel$
\end{enumerate}
\label{def:abstraction_relation}
\end{definition}

The relation $q_i~\crel~q_j$ is a sort of \textit{conjectural consequence 
relation}~\cite{Flach96} that allows us to conjecture the presence of $q_j$ 
from the observation of $q_i$. The transitive closure of the abstraction 
relation is a strict partial order relation between the domain observables, such 
that $q_i < q_j \Leftrightarrow q_i ~\crel^+ \!q_j$; that is, if and only if 
$\exists q_{k_0},\ldots,q_{k_n}\in \mathcal{Q}$ such that $q_{k_0}=q_i$, 
$q_{k_n}=q_j$ and for all $m$, with $0\leq m < n$, it holds that $q_{k_m} 
~\crel~ q_{k_{m+1}}$. We denote by $q_i=q_{k_0}~\crel~ q_{k_1}~\crel~ 
\ldots~\crel~ q_{k_n}=q_j$ an {\em abstraction sequence} in n steps that allows 
the conjecture of $q_j$ from $q_i$. This order relation defines an abstraction 
hierarchy among the observables in $\mathcal{Q}$. From the definition of a 
strict partial order, there must be at the base of this hierarchy at least one 
observable we call $q_0$, corresponding in the domain of electrocardiography to 
the digital signal.

\begin{example}
{\em  Let $\mathcal{Q} = \{q_{Pw}, q_{QRS}, q_{Tw}, q_{N}, q_{V}, q_{VB}\}$ and
$\mathcal{G}= \{G_N, G_{VB}\}$, containing the knowledge represented in
examples~\ref{ex:ncy} and~\ref{ex:vbigeminy}. The derived abstraction relation
states that $q_{Pw}, q_{QRS}, q_{Tw}~ \crel ~q_{N}$, and $q_{N}, q_{V}~ \crel~
q_{VB}$. Intuitively, we can see that this relation splits the observables into 
three abstraction levels: the wave level, describing the activation/recovery of 
the different heart chambers; the heartbeat level, describing each cardiac cycle 
by its origin in the muscle tissue; and the rhythm level, describing the 
dynamic behavior of the heart over multiple cardiac cycles. These levels match 
those commonly used by experts in electrocardiogram analysis~\cite{Marriott08}.}
\end{example}

It is worth noting that the abstraction relation is only established between 
observables in the $A_P$ set. This provides flexibility in defining the evidence 
forming the context of a pattern, as this may belong to different abstraction 
levels.

\begin{definition}
 We define an \textbf{abstraction model} as a tuple $\mathcal{M} = \langle
\mathcal{Q}, \crel, \mathcal{G} \rangle$, where $\mathcal{Q}$ is the set of
domain observables, $\crel$ is an abstraction relation between such observables,
and $\mathcal{G}$ is the available knowledge as a set of abstraction grammars.
\end{definition}

The successive application of the available abstraction grammars results in a 
series of observations organized in a hierarchy of abstraction, according to the 
order relation between observables as described above. We are able to define an 
interpretation problem as follows.

\begin{definition}
 We define an \textbf{interpretation problem} as a pair $IP=\langle\mathcal{O}, 
\mathcal{M}\rangle$, where $\mathcal{O}= (o_1,o_2,\ldots,o_i,\ldots)$ is a 
sequence of observations requiring interpretation and $\mathcal{M}$ is an 
abstraction model of the domain.
\end{definition}

It is worth mentioning that this definition of an abductive interpretation 
problem differs from the common definition of an abductive diagnosis problem, 
where the difference between normal and faulty behaviors is explicit, leading to 
the role of faulty manifestations that guide the abductive process of diagnosis. 
In contrast, in the present framework all the observations have the same status, 
and the objective of the interpretation process is to provide an interpretation 
of what is observed at the highest possible abstraction level in terms of the 
underlying processes. As we will see later, some observables may stand out 
amongst others regarding the efficiency of the interpretation process, as 
salient features that can draw some sort of perceptual attention. 

As discussed above, any observable $q\in Q_P$ can appear multiple times as 
different pieces of evidence for an abstraction pattern $P$, in the form of 
findings collected in the set $M_P$. As a consequence, $P$ can predict multiple 
observations of the set $\mathcal{O}$ for a given observable $q\in Q_P$, each 
of these corresponding to one of the findings of the set $M_P$ through a 
matching relation. This matching relation is a matter of choice for the agent in 
charge of the interpretation task, by selecting from the evidence the 
observation corresponding to each finding in a given pattern.

\begin{definition}
Given an interpretation problem $IP$, a \textbf{matching relation} for a pattern 
$P\in \mathcal{P}$ is an injective relation in $ M_P \times \mathcal{O}$, 
defined by  $m^q \leftarrowtail o$ if and only if $o=\langle 
q,\mathbf{v},t^b,t^e\rangle \in O(q)\subseteq \mathcal{O}$ and $m^q=\langle 
\psi,\mathbf{A},T^b, T^e\rangle \in M_P$, such that 
$(A_1=v_1,\ldots,A_{n_q}=v_{n_q})$, $T^b=t^b$ and $T^e=t^e$.
\end{definition}

A matching relation makes an assignment of a set of observations to a set of 
findings of a certain pattern, leading us to understand the interpretation 
problem as a search within the available evidence for a valid assignment for the 
constraints represented in an abstraction pattern.

From the notion of matching relation we can design a mechanism for abductively 
interpreting a subset of observations in $\mathcal{O}$ through the use of 
abstraction patterns. Thus, a matching relation for a given pattern allows us to 
hypothesize new observations from previous ones, and to iteratively incorporate 
new evidence into the interpretation by means of a hypothesize-and-test cycle. 
The notion of abstraction hypothesis defines those conditions that a subset of 
observations must satisfy in order to be abstracted by a new observation, and 
makes it possible to incrementally build an interpretation from the 
incorporation of new evidence.

\begin{definition}
 Given an interpretation problem $IP$, we define an \textbf{abstraction
hypothesis} as a tuple $\hbar=\langle o_h,P,\leftarrowtail\rangle$, where
$P=\langle h,M_{P},C_{P},\varTheta_{P}\rangle\in \mathcal{P}$,
$\leftarrowtail \subseteq M_P \times \mathcal{O}$, and we denote
$O_{\hbar}=codomain(\leftarrowtail)$, satisfying: 
\begin{enumerate}
\item $o_h\in O(h)$.
\item $o_h=\varTheta_P(O_{\hbar})$.
\item $C_P(\mathbf{A}_h,T^b_h,T^e_h,\mathbf{A}_1,T^b_1,T^e_1,\ldots,$ 
$\mathbf{A}_n,T^b_n,T^e_n)\vert _{o_h, o_1,\ldots,o_n\in O_{\hbar}}$ is 
satisfied.
\end{enumerate}
\label{def:abstraction_hypothesis}
\end{definition}

These conditions entail: (1) an abstraction hypothesis guesses an observation of 
the observable hypothesized by the pattern; (2) a new observation is obtained 
from the application of the observation procedure to those observations being 
assigned to the set of findings $M_P$ by the matching relation; and (3) the 
observations taking part in an abstraction hypothesis must satisfy those 
constraints of the pattern whose variables are assigned a value by the 
observations.

Even though the matching relation is a matter of choice, and therefore a 
conjecture in itself, some additional constraints may be considered as default 
assumptions. An important default assumption in the abstraction of a periodic 
process states that consecutive observations are related by taking part in the 
same hypothesis, defining the basic period of the process. This assumption 
functions as a sort of operative hypothesis of the abstraction task:

\begin{defassumption}
 \emph(Basic periodicity) Periodic findings in an abstraction pattern must be
assigned consecutive observations by any matching relation: 
\begin{displaymath}
 \forall m^q_i,m^q_{i+1}\in M^q_P, m^q_i \leftarrowtail o_j \wedge
\mathrm{q-}\mathtt{succ}(o_j)\in O_{\hbar} \Rightarrow m^q_{i+1} \leftarrowtail
\mathrm{q-}\mathtt{succ}(o_j)
\end{displaymath}
\end{defassumption}

\begin{figure}[h!]
 \centering
 \includegraphics[width=\textwidth]{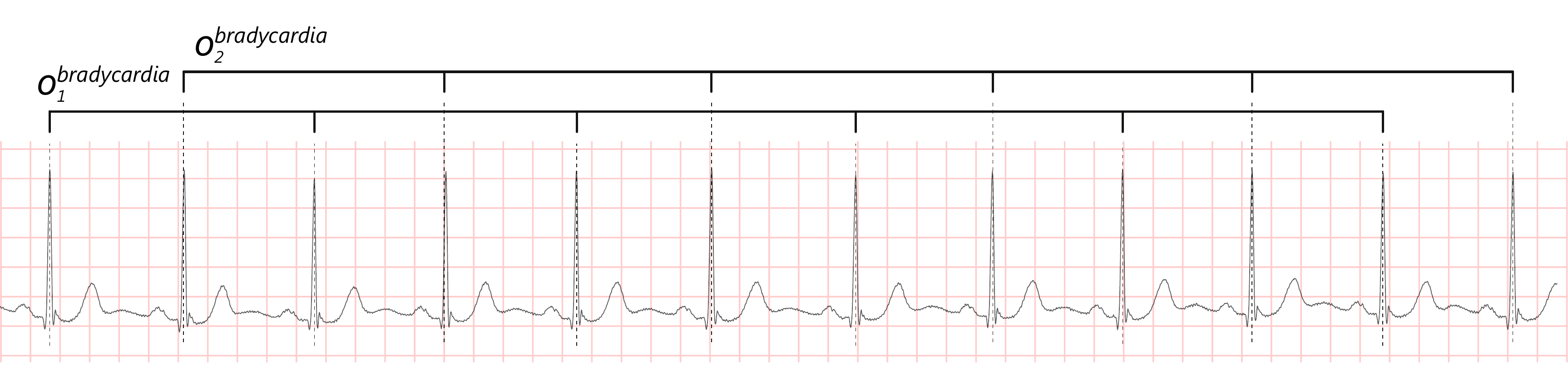}
 \caption{Motivation for the assumption of basic periodicity. [Source: MIT-BIH 
arrhythmia DB, recording: 103, between 00:40.700 and 00:51.200]}
 \label{fig:defassumpt2_motiv}
\end{figure}

This default assumption allows us to avoid certain combinations of abstraction 
hypotheses that, although formally correct, are meaningless from an 
interpretation point of view. For example, without the assumption of basic 
periodicity, a normal rhythm fragment might be abstracted by two alternating 
bradycardia hypotheses, as shown in Figure~\ref{fig:defassumpt2_motiv}.

The set of observations that may be abstracted in an interpretation problem $IP$ 
is $O(domain(\crel))$, that is, observations corresponding to observables 
involved in the set of findings to be abstracted by some abstraction pattern. An 
abstraction hypothesis defines in the set of observations $\mathcal{O}$ a 
counterpart of the subsets $A_{P}$ and $E_{P}$ of the set of findings $M_{P}$ of 
a pattern $P$, resulting from the selection of a set of observations $O_{\hbar} 
\subseteq \mathcal{O}$ by means of a matching relation, satisfying those 
requirements shown in the definition \ref{def:abstraction_hypothesis}.

\begin{definition}
 Given an interpretation problem $IP$ and an abstraction hypothesis 
$\hbar=\langle o_h,P,\leftarrowtail\rangle$, we define the following sets of
observations: 
\begin{itemize}
 \item $\mathtt{abstracted\_by}(o_h) = \{o \in O_{\hbar} ~|~  m_i^q 
\leftarrowtail o \wedge m_i^q \in A_P\}$. 
 \item $\mathtt{environment\_of}(o_h) = \{o \in O_{\hbar} ~|~ m_i^q 
\leftarrowtail o \wedge m_i^q \in E_P\}$.
 \item $\mathtt{evidence\_of}(o_h) = \mathtt{abstracted\_by}(o_h) \cup
\mathtt{environment\_of}(o_h)$. 
\end{itemize}
\end{definition}

We denote by $\mathtt{abstracted\_by}(o_h)$ the set of observations abstracted
by $o_h$ and which are somehow its constituents, while
$\mathtt{environment\_of}(o_h)$ denotes the evidential context of $o_h$.  We
denote by $\mathtt{evidence\_of}(o_h)$ the set of all observations supporting a
specific hypothesis. Since the matching relation is injective, it follows that
$\mathtt{abstracted\_by}(o_h) \cap \mathtt{environment\_of}(o_h) = \varnothing$.

The definition of these sets can be generalized to include as arguments a set of
observations $O=\{o_{h_1},...,o_{h_m}\}$ from a set of abstraction hypotheses
$\hbar _1,...,\hbar _m$:

\begin{itemize}
 \item $\mathtt{abstracted\_by}(O) = \bigcup _{o_h\in O}
\mathtt{abstracted\_by}(o_h)$ 
 \item $\mathtt{environment\_of}(O) = \bigcup
_{o_h\in O} \mathtt{environment\_of}(o_h)$.
 \item $\mathtt{evidence\_of}(O) = \bigcup _{o_h\in O}
\mathtt{evidence\_of}(o_h)$. 
\end{itemize}

As a result of an abstraction hypothesis, a new observation $o_h$ is generated 
which can be included in the set of domain observations, so that $\mathcal{O}= 
\mathcal{O} \cup \{o_h\}$. In this way, an interpretation can be incrementally 
built from the observations, by means of the aggregation of abstraction 
hypotheses.

\begin{definition}
 Given an interpretation problem $IP$, an \textbf{interpretation} $I$ is 
defined as a set of abstraction hypotheses $\{\hbar _1,\ldots,\hbar _m\}$.
\end{definition}

An interpretation can be rewritten as $I=\langle O_{I}, P_I, 
\leftarrowtail_I\rangle$, where $O_I=\{o_{h_1},\ldots$ $,o_{h_m}\}$ is the set 
of observations guessed by performing multiple abstraction hypotheses; 
$P_I=\{P_1,\ldots,P_m\}$ is the set of abstraction patterns used in the 
interpretation; and $\leftarrowtail_I = \leftarrowtail_{\hbar _1} \cup \ldots 
\cup \leftarrowtail _{\hbar_m}$  $\subseteq (M_1\cup \ldots \cup M_m) \times 
\mathcal{O}$ is the global matching relation. It should be noted that the global 
matching relation $\leftarrowtail_I$ is not necessarily injective, since some 
observations may simultaneously belong to both the $\mathtt{abstracted\_by}()$ 
and $\mathtt{environment\_of}()$ sets of different observations.

From a given interpretation problem $IP$, multiple interpretations can be 
abductively proposed through different sets of abstraction hypotheses. Indeed, 
the definition of interpretation is actually weak, since even an empty set $I = 
\varnothing$ is formally a valid interpretation. Thus, we need additional 
criteria in order to select the solution to the interpretation problem as the 
best choice among different possibilities \cite{Peng90}.

\begin{definition}
 Given an interpretation problem $IP$, an interpretation $I$ is a \textbf{cover}
of $IP$ if the set of observations to be interpreted  $O(domain(\crel))\subseteq
\mathcal{O}$ is included in the set of observations abstracted by $I$, that is,
$O(domain(\crel))\subseteq \mathtt{abstracted\_by}(O_I)$.
\end{definition}

\begin{definition}
 Given an interpretation problem $IP$, two different abstraction hypotheses 
$\hbar$ and $\hbar'$ of the mutually exclusive observables $q_h$ and $q_{h'}$ 
are \textbf{alternative hypotheses} if and only if $\mathtt{abstracted\_by}(o_h) 
\cap \mathtt{abstracted\_by}(o_{h'}) \neq \varnothing$.
\end{definition}

\begin{example}
{\em A \textit{ventricular trigeminy} is an infrequent arrhythmia very similar 
to \textit{ventricular bigeminy}, except that the ectopic heart beats occur 
after every pair of normal beats instead of after each one. The grammar for 
hypothesizing a ventricular trigeminy $q_{VT}$ would therefore be very similar 
to that described in example~\ref{ex:vbigeminy}, with the difference that each 
$q_V$ finding would appear after every pair of $q_N$ findings. These two 
processes are mutually exclusive, insofar as the heart can develop just one of 
these activation patterns at a given time. For this reason, in the event of an 
observation of $q_V$, this may be abstracted by either a $q_{VB}$ or a $q_{VT}$ 
hypothesis, but never by both simultaneously.}
\end{example}

\begin{definition}
 Given an interpretation problem $IP$, a cover $I$ for $IP$ is
\textbf{exclusive} if and only if it contains no alternative hypotheses. 
\end{definition}

Thus, two or more different hypotheses of mutually exclusive observables
abstracted from the same observation will be incompatible in the same
interpretation, since inferring both a statement and its negation is logically
prevented, and therefore only one of them can be selected.

On the other hand, a parsimony criterion is required, in order to disambiguate 
the possible interpretations to select as the most plausible those of which the 
complexity is minimum \cite{Peng90}. We translate this minimum complexity in 
terms of minimal cardinality. 

\begin{definition}
Given an interpretation problem $IP$, a cover $I$ for $IP$ is \textbf{minimal},
if and only if its cardinality is the smallest among all covers for $IP$. 
\end{definition}

Minimality introduces a parsimony criterion on hypothesis generation, promoting 
temporally maximal hypotheses, that is, those hypotheses of a larger scope 
rather than multiple equivalent hypotheses of smaller scope. For example, 
consider an abstraction pattern that allows the conjecture of a regular cardiac 
rhythm from the presence of three or more consecutive heart beats. Without a 
parsimony criterion, a sequence of nine consecutive beats could be abstracted by 
up to three consecutive rhythm observations, even when a single rhythm 
observation would be sufficient and better.

\begin{definition}
The \textbf{solution} of an interpretation problem $IP$ is the set of all
minimal and exclusive covers of $IP$.
\end{definition}

This definition of solution is very conservative and has limited practical 
value, since the usual objective is to obtain a small set of interpretations 
explaining what has been observed (and ideally only a single one). However, it 
allows us to characterize the problem in terms of complexity. Abduction has been 
formulated under different frameworks according to the task to be addressed, but 
has always been found an intractable problem in the general case 
\cite{Josephson94}. The next theorem proves that an interpretation problem is 
also an intractable problem.

\begin{theorem}
 Finding the solution to an interpretation problem is NP-hard.
\end{theorem}

\noindent \textit{Proof:} We will provide a polynomial-time reduction of the 
well-known set covering problem to an interpretation problem. Given a set of 
elements $U=\{u_1,\ldots,u_m\}$ and a set $S$ of subsets of $U$, a cover is a 
set $C\subseteq S$ of subsets of $S$ whose union is $U$. In terms of complexity 
analysis, two different problems of interest are identified:

\begin{itemize}
 \item A set covering decision problem, stating that given a pair $(U,S)$ and an
integer $k$ the question is whether there is a set covering of size $k$ or less.
This decision version of set covering is NP-complete.
 \item A set covering optimization problem, stating that given a pair $(U,S)$
the task is to find a set covering that uses the fewest sets. This optimization
version of set covering is NP-hard.
\end{itemize}
We will therefore reduce the set covering problem to an interpretation problem
by means of a polynomial-time function $\varphi$. Thus, we shall prove that
$\varphi(U,S)$ is an interpretation problem, and there is a set covering of
$\varphi(U,S)$ of size $k$ or less if and only if there is a set covering of $U$
in $S$ of size $k$ or less.

Given a pair $(U,S)$, let $\varphi (U,S)=\langle\mathcal{O}, \mathcal{M}\rangle$
where:
\begin{enumerate}
 \item $\mathcal{O}=U=\{u_1,\ldots,u_m\}$, such that $u_i=\langle 
q,true,i\rangle$ and $q=\langle \psi,present,T\rangle$.
 \item $\mathcal{M}=\langle Q, \crel,\mathcal{P}\rangle$, such that 
$domain(\crel)\!=q$.
 \item $\forall s=\{u_{i_1},\ldots,u_{i_n}\}\in S,~ \exists P\in \mathcal{P}$,
being $P=\langle q_P,M_P,C_P,\varTheta_P\rangle$, where:
\begin{itemize}
\item $q\ \crel\ q_P$ and $P\neq P' \Rightarrow q_P \neq q_{P'}$.
\item $M_P=A_P=M^q_P=\{m^q_1=\langle \psi, present_1,T_1\rangle,\ldots,m^q_n\}$.
\item $C_P= \{\bigwedge ^n_{k=1}T_k=k; T^b_h=min 
\{T_k\};T^e_h=max \{T_k\}\}$.
\item $present_P=\varTheta_P(m^q_1,\ldots,m^q_n)=\bigwedge 
^n_{k=1}present_k$.
\end{itemize}
\end{enumerate}
Thus, $\varphi (U,S)$ is an interpretation problem according to this definition.
On the other hand, $\varphi (U,S)$ can be built in polynomial time. In addition,
for all $s\in S$ there exists an abstraction hypothesis $\hbar=\langle o_h,P,
\leftarrowtail \rangle$ such that:
\begin{enumerate}
\item $o_h=\langle h, true, min_{u_i \in s} 
\{i\},max_{u_i \in s} \{i\}\rangle$.
\item $u_i\in s \Rightarrow u_i\in codomain(\leftarrowtail)$.
\item $\leftarrowtail$ provides a valid assignment, since the set of 
observations satisfying $\varTheta_P=true$ also satisfies the constraints in 
$C_P$.
\end{enumerate}

Since each abstraction hypothesis involves a different abstraction pattern there 
are no alternative hypotheses in any interpretation of $\varphi(U,S)$.

Suppose there is a set covering $C\subseteq S$ of $U$ of size $k$ or less. For 
all $u\in U$ there exists $c_i \in C-\{\varnothing\}$ such that $u\in c_i$ and, 
by the above construction, there exists $\hbar_i \in I$ such that 
$\mathtt{abstracted\_by}(o_{h_i})=\{u\in codomain(\leftarrowtail 
_{\hbar_i})\}=\{u\in c_i\}=c_i$, and therefore, $O(domain(\crel))\subseteq 
\bigcup _{\hbar _i\in I}\mathtt{abstracted\_by}(o_{h_i})=\bigcup _{i} c_i = C$.
That is, the set of abstraction hypotheses $I$ is an exclusive cover of the 
interpretation problem $\varphi(U,S)$ of size $k$ or less.

Following the same reasoning as for the set covering optimization problem, 
finding a minimal and a exclusive cover of an interpretation problem 
$\varphi(U,S)$ is NP-hard, since we can use the solution of this problem to 
check whether there is an exclusive cover of the interpretation problem of size 
$k$ or less, and this has been proven above to be NP-complete. $\Box$

\section{Solving an interpretation problem: A heuristic search approach}
\label{sec:search}
The solution set for an interpretation problem $IP$ consists of all exclusive 
covers of $IP$ having the minimum possible number of abstraction hypotheses. 
Obtaining this solution set can be stated as a search on the set of 
interpretations of $IP$. The major source of complexity of searching for a 
solution is the local selection, from the available evidence in $\mathcal{O}$, 
of the most appropriate matching relation for a number of abstraction hypotheses 
that can globally shape a minimal and exclusive cover of $IP$. 

Nevertheless, the whole concept of solution must be revised in practical terms, 
due to the intractability of the task and the incompleteness of the abstraction 
model, that is, of the available knowledge. Indeed, we assume that any 
realistic abstraction model can hardly provide a cover for every possible 
interpretation problem. Hence the objective should shift from searching for a 
solution to searching for an approximate solution. 

Certain principles applicable to the interpretation problem can be exploited in
order to approach a solution in an iterative way, bounding the combinatorial
complexity of the search. These principles can be stated as a set of heuristics
that make it possible to evaluate and discriminate some interpretations against
others from the same base evidence: 
\begin{itemize}
\item A \textit{coverage principle}, which states the preference for
interpretations explaining more initial observations.
\item A \textit{simplicity principle}, which states the preference for 
interpretations with fewer abstraction hypotheses.
\item An \textit{abstraction principle}, which states the
preference for interpretations involving higher abstraction levels.
\item A \textit{predictability principle}, which states the preference for
interpretations that properly predict future evidence. 
\end{itemize}

The coverage and simplicity principles are used to define a cost measure for the 
heuristic search process~\cite{Edelkamp2011}, while the abstraction and 
predictability principles are used to guide the reasoning process, in an attempt 
to emulate the same shortcuts used by humans. 

Given an interpretation problem $IP$, a heuristic vector for a certain
interpretation $I$ can be defined to guide the search, as $\epsilon(I)=(1 -
\varsigma(I), \kappa(I))$, where
$\varsigma(I)=|\mathtt{abstracted\_by}(O_{I})|/|O(domain(\crel))|$ is the {\em
covering ratio} of $I$, and $\kappa(I) = |O_I|$ is the {\em complexity} of $I$.
The main goal of the search strategy is to approach a solution with a maximum
covering ratio and a minimum complexity, which is equivalent to the minimization
of the heuristic vector. The covering ratio will be considered the primary
heuristic, and complexity will be considered for ranking interpretations
with the same covering ratio. The $\epsilon(I)$ heuristic is intuitive and very
easy to calculate, but as a counterpart it is a non-admissible heuristic, since
it is not monotone and may underestimate or overestimate the true
goal covering. Therefore optimality cannot be guaranteed and we require an
algorithm efficient with this type of heuristic. We propose the
\begin{small}CONSTRUE()\end{small} algorithm, whose pseudocode is shown in
Algorithm~\ref{alg:construe}. This algorithm is a minor variation of the K-Best
First Search algorithm~\cite{Edelkamp2011}, with partial expansion to reduce the
number of explored nodes.

\begin{algorithm}[h!]
\captionsetup{font=small}
\caption{CONSTRUE search algorithm.}
\label{alg:construe}
\begin{algorithmic} [1]
  \Function{CONSTRUE}{$IP$}
    \State \textbf{var} $I_0 = \varnothing$
    \State \textbf{var} $K = max(|\{q_j \in \mathcal{Q} ~|~ q_i~\crel~q_j, q_i
\in \mathcal{Q}\}|)$
    \State \Call{set\_focus}{$I_0, o_1$}    
    \State \textbf{var} $open =$ \Call{sorted}{$[\langle\epsilon(I_0),
I_0\rangle]$}    
    \State \textbf{var} $closed =$ \Call{sorted}{$[]$}
    \While{$open \neq \varnothing$}      
      \ForAll{$I \in open[0\ldots K]$} \label{ln:kselect}
        \State $I' = $ \Call{next}{\Call{get\_descendants}{$I$}}
        \If{$I'$ is null}
          \State $open = open - \{\langle \epsilon(I),I\rangle\}$
          \State $closed = closed \cup \{\langle \epsilon(I), I\rangle\}$
\label{ln:toclosed}
        \ElsIf{$\varsigma(I') = 1.0$} \label{ln:solution}
          \State \Return $I'$
        \Else
          \State $open = open \cup \{\langle \epsilon(I'),I'\rangle\}$
      \EndIf
      \EndFor      
    \EndWhile
    \State \Return $min(closed)$ \label{ln:minsolution}
  \EndFunction  
\end{algorithmic}
\end{algorithm}

The \begin{small}CONSTRUE()\end{small} algorithm takes as its input an 
interpretation problem $IP$, and returns the first interpretation found with 
full coverage, or the interpretation with the maximum covering ratio and minimum 
complexity if no covers are found, using the abstraction and predictability 
principles in the searching process. To do this, it manages two ordered 
lists of interpretations, named $open$ and $closed$. Each interpretation is 
annotated with the computed values of the heuristic vector. The $open$ list 
contains those partial interpretations that can further evolve by (1) appending 
new hypotheses or (2) extending previously conjectured hypotheses to subsume or 
predict new evidence. This $open$ list is initialized with the trivial 
interpretation $I_0=\varnothing$. The $closed$ list contains those 
interpretations that cannot explain more evidence.

At each iteration, the algorithm selects the $K$ most promising interpretations 
according to the heuristic vector (line~\ref{ln:kselect}), and partially expands 
each one of them to obtain the next descendant node $I'$. If this node is a 
solution, then the process ends by returning it (line~\ref{ln:solution}), 
otherwise it is added to the $open$ list. The partial expansion ensures that the 
$open$ list grows at each iteration by at most $K$ new nodes, in order to save 
memory. When a node cannot expand further, it is added to the $closed$ list 
(line~\ref{ln:toclosed}), from which the solution is taken if no full coverages 
are found (line~\ref{ln:minsolution}).

The selection of a value for the $K$ parameter depends on the problem at hand.
We select its value as $K = max(|\{q_j \in \mathcal{Q} ~|~ q_i ~\crel~ q_j, q_i
\in \mathcal{Q}\}|)$, that is, as the maximum number of observables that can be
abstracted from any observable $q_i$. The intuition behind this choice is that
at any point in the interpretation process, and with the same heuristic values,
the same chance is given to any plausible abstraction hypothesis in order to
explain a certain observation.

In order to expand the current set of interpretations, the 
\begin{small}GET\_DESCEND-ANTS()\end{small} function relies on different 
reasoning modes, that is, different forms of abduction and deduction, which are 
brought into play under the guidance of an attentional mechanism. Since 
searching for a solution finally involves the election of a matching relation, 
both observations and findings should be included in the scope of this 
mechanism. Hence, a focus of attention can be defined to answer the following 
question: which is the next observation or finding to be processed? The answer 
to this question takes the form of a hypothesize-and-test cycle: if the 
attention focuses on an observation, then an abstraction hypothesis explaining 
this observation should be generated (hypothesize); however, if the attention 
focuses on a finding predicted by some hypothesis, an observation should be 
sought to match such finding (test). Thus, the interpretation problem is solved 
by a reasoning strategy that progresses incrementally over time, coping with new 
evidence through the dynamic generation of abstraction patterns from a finite 
number of abstraction grammars, and bounding the theoretical complexity by a 
parsimony criterion.

To illustrate and motivate the reasoning modes implemented in building
interpretations and supporting the execution of the
\begin{small}CONSTRUE()\end{small} algorithm, we use a simple, but complete,
interpretation problem.

\begin{example}
\label{ex:interp_problem}
{\em Let $\mathcal{Q} = \{q_{wave}, q_{Pw}, q_{QRS}, q_{Tw}, q_N\}, \mathcal{G}
= \{G_w,G_N, G_{Tw}\}$, where $G_w$ models the example~\ref{ex:wave}, $G_N$ is
described in example~\ref{ex:ncy}, and $G_{Tw}=(\{H, D\},\{q_{QRS}, q_{wave}\}, 
H, R)$ describes the knowledge to conjecture a T wave with the following rules:

\begin{align*}
H &\rightarrow q_{QRS} D & & {\scriptstyle \{P_H :=~ \langle q_{Tw}, 
 M_H=\varnothing, C_H = \varnothing,\varTheta_H= \varnothing \rangle,}\\
 & & &~ {\scriptstyle C_{QRS} :=~ \{ 80ms \leq T^b_{Tw}-T^e_{QRS} \leq 
 120ms;~ T^e_{Tw}-T^b_{QRS} \leq 520ms \},}\\
 & & &~ {\scriptstyle A_{QRS} :=~ false,}\\
 & & &~ {\scriptstyle P_D :=~ \langle q_{Tw}, M_D=\{m^{QRS}\}, C_D=C_{QRS}, 
 \varTheta_D = \varnothing \rangle}\\[-5pt]
 & & & \scriptstyle\}\\
D &\rightarrow q_{wave} & & {\scriptstyle \{ P_D :=~ \langle q_{Tw}, 
 M_D=\{m^{QRS}\}, C_D=C_{QRS}, \varTheta_D = \varnothing \rangle,}\\
 & & &~ {\scriptstyle C_{wave} :=~ \{ T^b_{Tw} = T^b_{wave};~ T^e_{Tw} = 
 T^e_{wave};~ max(\text{\textit{diff}}(sig[m^{wave}]) \leq 0.7 \cdot 
 max(\text{\textit{diff}}(sig[m^{QRS}]))\},}\\
 & & &~ {\scriptstyle A_{wave} :=~ true,}\\
 & & &~ {\scriptstyle P :=~ \langle q_{Tw}, M_P=M_D\cup\{m^{wave}\}, C_P= C_D 
\cup C_{wave}, \varTheta_P = \mathtt{Tw\_delin}(T^b_{QRS},T^e_{QRS},T^b_{wave}, 
T^e_{wave}) \rangle}\\[-5pt]
 & & &\scriptstyle\}
\end{align*}

This grammar hypothesizes the observation of a T wave from a wave appearing 
shortly after the observation of a QRS complex, requiring a significant decrease 
in the maximum slope of the signal (in the constraint definition $C_{wave}$, the 
expression ``$max(\text{\textit{diff}}(sig[m])$'' stands for the maximum 
absolute value of the derivative of the ECG signal between $T^b_m$ and $T^e_m$). 
The observation procedure of the generated pattern is denoted as 
$\mathtt{Tw\_delin()}$, and may be any of the methods described in the 
literature for the delineation of T waves, such as in~\cite{Laguna94}.

In addition to the $P_{wave}$ pattern generated by $G_w$ and detailed in
example~\ref{ex:wave}, $G_N$ and $G_{Tw}$ generate the following abstraction
patterns:

\begin{align*}
P_N &= \langle q_N, A_{P_N} = \{m^{Pw}, m^{QRS}, m^{Tw}\} \cup E_{P_N} =
\varnothing, C_{P_N}, \varTheta_{P_N}=\varnothing\rangle\\
P_{Tw} &= \langle q_{Tw}, A_{P_{Tw}} = \{m^{wave}\} \cup E_{P_{Tw}} =
\{m^{QRS}\}, C_{QRS} \cup C_{wave}, \mathtt{Tw\_delin()} \rangle 
\end{align*}

Finally, let $\mathcal{O} = \{o^{wave}_1 = \langle q_{wave}, \varnothing, 0.300, 
0.403\rangle, o^{wave}_2 = \langle q_{wave}, \varnothing, 0.463,$ $0.549 
\rangle, o^{Pw} = \langle q_{Pw}, \varnothing, 0.300, 0.403\rangle, o^{QRS} = 
\langle q_{QRS}, \varnothing, 0.463, 0.549 \rangle\}$ be a set of initial 
observations including a P wave and a QRS complex abstracting two wave 
observations located at specific time points.

Given this interpretation problem, Figure~\ref{fig:full_interpretation} shows 
the starting point for the interpretation, where the root of the interpretation 
process is the trivial interpretation $I_0$, and the attention is focused on 
the first observation. The sequence of reasoning steps towards the resolution 
of this interpretation problem will be explained in the following subsections.}
\end{example}

\subsection{Focus of attention}

The focus of attention is modeled as a stack; thus, once the focus is set on a 
particular observation (or finding), any observation that was previously under 
focus will not return to be focused on until the reasoning process on the 
current observation is finished. Algorithm~\ref{alg:interpretation_desc} shows 
how the different reasoning modes are invoked based on the content of the focus 
of attention, resulting in a hypothesize-and-test cycle.

\begin{algorithm}
\captionsetup{font=small}
\caption{Method for obtaining the descendants of an interpretation using
different reasoning modes based on the content of the focus of attention.}
\label{alg:interpretation_desc}
\begin{algorithmic} [1]
  \Function{get\_descendants}{$I$}
    \State \textbf{var} $focus =$ \Call{get\_focus}{$I$}.\Call{top}{$ $}
    \State \textbf{var} $desc = \varnothing$
    \If{\Call{is\_observation}{$focus$}} \label{ln:obsin}
      \If{$focus=o_h ~|~ \hbar \in I$}
        \State $desc = $ \Call{deduce}{$I, focus$}
      \EndIf
      \State $desc = desc ~\cup $ \Call{abduce}{$I, focus$} $\cup$
\Call{advance}{$I, focus$} \label{ln:obsout}
    \ElsIf{\Call{is\_finding}{$focus$}}
      \State $desc = $ \Call{subsume}{$I, focus$} $\cup$ \Call{predict}{$I,
focus$} \label{ln:finding}    
    \EndIf
    \State \Return $desc$    
  \EndFunction
\end{algorithmic}
\end{algorithm}

Lines~\ref{ln:obsin}-\ref{ln:obsout} generate the descendants of an 
interpretation $I$ when there is an observation at the top of the stack. These 
descendants are the result of two possible reasoning modes: the deduction of new 
findings, performed by the \begin{small}DEDUCE()\end{small} function, provided 
that the observation being focused on is an abstraction hypothesis; and the 
abduction of a new hypothesis explaining the observation being focused on, 
performed by the \begin{small}ABDUCE()\end{small} function. A last descendant is 
obtained using the \begin{small}ADVANCE()\end{small} function, which simply 
restores the previous focus of attention by means of a 
\begin{small}POP()\end{small} operation. If the focus is then empty, 
\begin{small}ADVANCE()\end{small} inserts the next observation to explain, 
which may be selected by temporal order in the general case, or by some 
domain-dependent saliency criterion to prioritize certain observations over 
others. By removing the observation at the top of the focus of attention, the 
\begin{small}ADVANCE()\end{small} function sets aside that observation as 
unintelligible in the current interpretation, according to the available 
knowledge.

If the top of the stack contains a finding, then 
Algorithm~\ref{alg:interpretation_desc} obtains the descendants of the 
interpretation from the \begin{small}SUBSUME()\end{small} and 
\begin{small}PREDICT()\end{small} functions (line~\ref{ln:finding}). The first 
of these functions looks for an existing observation satisfying the constraints 
on the finding focused on, while the second makes predictions about observables 
that have not yet been observed. All of these reasoning modes are described 
separately and detailed below; we will illustrate how the 
\begin{small}CONSTRUE()\end{small} algorithm combines these in order to solve 
the interpretation problem in Example~\ref{ex:interp_problem}.

\subsection{Building an interpretation: Abduction}

Algorithm~\ref{alg:abduction} enables the abductive generation of new 
abstraction hypotheses. It is applied when the attention is focused on an 
observation that can be abstracted by some abstraction pattern, producing a new 
observation at a higher level of abstraction.

\begin{algorithm}
\captionsetup{font=small}
\caption{Moving forward an interpretation through abduction.}
\label{alg:abduction}
\begin{algorithmic} [1]
\Function{abduce}{$I, o_i$}
    \State \textbf{var} $desc = \varnothing$
    \ForAll{$G_h=\langle V_N, V_T, H, R \rangle \in \mathcal{G} ~|~ q(o_i) 
~\crel~ h$} \label{ln:abspat}
      \ForAll{$(U \rightarrow q V) \in R ~|~ q(o_i) ~\mathtt{is\_a}~ q \wedge
A_q = true $} \label{ln:absfind}
        \State $P_{V} = \langle h, M_{V}=\{m^q\}, C_{V}, \varTheta_{V} \rangle$ 
\label{ln:newpat}
        \State $\hbar = \langle o_h, P_{V}, \leftarrowtail_\hbar=\{m^q 
          \leftarrowtail o_i\} \rangle$ \label{ln:newobs}
        \State $L_\hbar = [(U \rightarrow q V)]; B_\hbar = U; E_\hbar = V$ 
\label{ln:linit}        
        \State $I' = \langle O_I \cup \{o_h\}, P_I \cup \{P_V\}, 
          \leftarrowtail_I \cup \leftarrowtail_\hbar \rangle$ \label{ln:hypin}
        \State $\mathcal{O} = \mathcal{O} \cup \{o_h\}$ \label{ln:hypout}
        \State \Call{get\_focus}{$I'$}.\Call{pop}{$ $} \label{ln:focuspop}
        \State \Call{get\_focus}{$I'$}.\Call{push}{$o_h$} \label{ln:focuspush}
        \State $desc = desc \cup \{I'\}$
      \EndFor      
    \EndFor    
    \State \Return $desc$
  \EndFunction
\end{algorithmic}
\end{algorithm}

The result of \begin{small}ABDUCE()\end{small} is a set of interpretations $I'$, 
each one adding a new abstraction hypothesis with respect to the parent 
interpretation $I$. To generate these hypotheses, we iterate through those 
grammars that can make a conjecture from the observation $o_i$ under focus 
(line~\ref{ln:abspat}). Then, for each grammar, each production including the 
corresponding observable $q(o_i)$ (line~\ref{ln:absfind}) initializes an 
abstraction pattern with a single finding of this observable 
(line~\ref{ln:newpat}), and a new hypothesis is conjectured with a matching 
relation involving both the observation under focus and the finding 
(line~\ref{ln:newobs}). A list structure $L_\hbar$ and two additional variables 
$B_\hbar$ and $E_\hbar$ are initialized to trace the sequence of productions 
used to generate the findings in the abstraction pattern; these will play an 
important role in subsequent reasoning steps (line~\ref{ln:linit}). Finally the 
new hypothesis opens a new interpretation (lines~\ref{ln:hypin}-\ref{ln:hypout}) 
focused on this hypothesis (line~\ref{ln:focuspush}).

In this way, the \begin{small}ABDUCE()\end{small} function implements, from a 
single piece of evidence, the hypothesize step of the hypothesize-and-test 
cycle. Below we explain the reasoning modes involved in the test step of the 
cycle.

\begin{example}
{\em
Let us consider the interpretation problem set out in 
example~\ref{ex:interp_problem} and the interpretation $I_0$ shown in 
Figure~\ref{fig:full_interpretation}. According to 
Algorithm~\ref{alg:interpretation_desc}, the \begin{small}ABDUCE()\end{small} 
function is used to move forward the interpretation, since the focus of 
attention points to an observation $o^{Pw}$. The abstraction pattern that 
supports this operation is $P_N$, and a matching relation is established with 
the $m^{Pw}$ finding. As a result, the following hypothesis is generated:
\begin{equation*}
 \hbar_1 = \langle o^N, P_N, \{m^{Pw} \leftarrowtail o^{Pw}\} \rangle
\end{equation*}

Figure~\ref{fig:full_interpretation} shows the result of this reasoning process,
in a new interpretation called $I_1$. Note that the focus of attention has been
moved to the newly created hypothesis 
(lines~\ref{ln:focuspop}-\ref{ln:focuspush} of the
\begin{small}ABDUCE()\end{small} function).
}
\end{example}

\subsection{Building an interpretation: Deduction}

This reasoning mode is applied when the attention is focused on an observation 
$o_h$ previously conjectured as part of an abstraction hypothesis $\hbar$ (see 
Algorithm~\ref{alg:deduction_h}). The \begin{small}DEDUCE()\end{small} function 
takes the evidence that has led to conjecture $o_h$ and tries to extend it with 
new findings which can be expected, i.e., deduced, from the abstraction 
grammar $G_h$ used to guess the observation. The key point is that this 
deduction process follows an iterative procedure, as the corresponding 
abstraction pattern is dynamically generated from the grammar. Hence the 
\begin{small}DEDUCE()\end{small} function aims to extend a partial matching 
relation by providing the next finding to be tested, as part of the test step of 
the hypothesize-and-test cycle.

\begin{algorithm}[t!]
\captionsetup{font=small}
\caption{Moving forward an interpretation through the deduction of new 
findings.} 
\label{alg:deduction_h}
\begin{algorithmic} [1]
\Function{deduce}{$I, o_h$}
    \State \textbf{var} $desc = \varnothing$        
    \If{$B_\hbar \neq H$} \label{ln:bin}
      \ForAll{$(X \rightarrow q B_\hbar) \in R$} \label{ln:prules}
        \State $P_{B_\hbar} = \langle h, M_{B_\hbar} = \{m^q\}, C_{B_\hbar}, 
\varTheta_{B_\hbar} \rangle$ \label{ln:newbpat}
        \ForAll{$(U \rightarrow q' V) \in L_\hbar$} \label{ln:travel_in}
          \State $P_V = \langle h, M_U \cup \{m^{q'}\}, C_U \cup C_V, 
\varTheta_V \rangle$
        \EndFor \label{ln:travel_end}        
        \State $\hbar = \langle o_h, P_{E_\hbar}, \leftarrowtail_\hbar \rangle$ 
\label{ln:bhypupdate}
        \State $I' = \langle O_I, P_I \cup \{P_{E_\hbar}\}, \leftarrowtail_I
\rangle$ \label{ln:bnewint}
        \State \Call{insert}{$L_\hbar, (X \rightarrow q B_\hbar), begin$}$; 
B_\hbar = X$ 
\label{ln:linsert}
        \State \Call{get\_focus}{$I'$}.\Call{push}{$m^q$} \label{ln:bpush}
        \State $desc = desc \cup \{I'\}$
      \EndFor \label{ln:bout}
    \Else \label{ln:ein}
      \ForAll{$(E_\hbar \rightarrow qX) \in R$} \label{ln:erules}
        \State $P_X = \langle h, M_{E_\hbar} \cup \{m^q\}, C_{E_\hbar} \cup C_X, 
\varTheta_X \rangle$
        \State $\hbar = \langle o_h, P_X, \leftarrowtail_\hbar \rangle$
        \State $I' = \langle O_I, P_I \setminus \{P_{E_\hbar}\}  \cup \{P_X\}, 
\leftarrowtail_I \rangle$
        \State \Call{insert}{$L_\hbar, (E_\hbar \rightarrow qX), end$}$; 
E_\hbar = X$
        \State \Call{get\_focus}{$I'$}.\Call{push}{$m^q$} 
\label{ln:ffocus}
        \State $desc = desc \cup \{I'\}$
      \EndFor \label{ln:eout}
    \EndIf    
    \State \Return $desc$
  \EndFunction
\end{algorithmic}
\end{algorithm}

Since the first finding leading to conjecture $o_h$ does not necessarily appear 
at the beginning of the grammar description, the corresponding abstraction 
pattern will not, in general, be generated incrementally from the first 
production of the grammar. Taking as a starting point the production used to 
conjecture $o_h$ (line~\ref{ln:absfind} in Algorithm \ref{alg:abduction}), the 
goal is to add a new finding by applying a new production at both sides, towards 
the beginning and the end of the grammar, using the information in the $L_\hbar$ 
list. The $B_\hbar$ variable represents the non-terminal at the left-hand side 
of the first production in $L_\hbar$, while $E_\hbar$ represents the 
non-terminal at the right-hand side of the last production in $L_\hbar$. Hence, 
this list has the form $L_\hbar = [(B_\hbar \rightarrow q'V'),(V' \rightarrow 
q''V''),\ldots,(V'^{n-1} \rightarrow q'^n E_\hbar)]$. In case $L_\hbar$ is 
empty, both variables $B_\hbar$ and $E_\hbar$ represent the $H$ non-terminal. 
With this information the sequence of findings supporting the hypothesis $\hbar$ 
can be updated in two opposite directions:

\begin{itemize}
 \item Towards the beginning of the grammar (lines~\ref{ln:bin}-\ref{ln:bout}): 
we explore the set of observables that may occur before the first finding 
according to the productions of the grammar (line~\ref{ln:prules}), and a new 
finding is deduced for each of these in different descendant interpretations. 
A new pattern $P_{B_\hbar}$ associated with the $B_\hbar$ non-terminal is 
initialized with the new finding (line~\ref{ln:newbpat}), and by moving along 
the sequence of productions generating the previous set of findings 
(lines~\ref{ln:travel_in}-\ref{ln:travel_end}) the pattern associated to the 
rightmost non-terminal $P_{E_\hbar}$ is updated with a new set of findings 
containing $m^q$. Consequently, the hypothesis and the interpretation are also 
updated (lines~\ref{ln:bhypupdate} and \ref{ln:bnewint}), and the applied 
production is inserted at the beginning of $L_\hbar$ (line~\ref{ln:linsert}). 
Finally the newly deduced finding is focused on (line~\ref{ln:bpush}).
  \item Towards the end of the grammar (lines~\ref{ln:ein}-\ref{ln:eout}): for 
each one of the observables that may occur after the last finding, a new finding 
$m^q$ is deduced, expanding the abstraction pattern associated with the new 
rightmost non-terminal $X$. After updating the hypothesis $\hbar$, the previous 
pattern $P_{E_\hbar}$ in the resulting interpretation $I'$ is replaced by 
the new one, $P_X$, and the applied production is inserted at the end of 
$L_\hbar$. Finally, the new finding is focused on (line~\ref{ln:ffocus}).
\end{itemize}

\begin{example}
{\em
Let us consider the interpretation problem set out in 
example~\ref{ex:interp_problem} and the interpretation $I_1$ shown in 
Figure~\ref{fig:full_interpretation}. Remember that the grammar used to generate 
the hypothesis in the focus of attention, $G_N$, has the following form:

\begin{align*}
H &\rightarrow q_{Pw}D \\
D &\rightarrow q_{QRS}E \\
E &\rightarrow q_{Tw} 
\end{align*}

In this situation, it is possible to deduce new findings from the $o^N$ 
hypothesis. Following Algorithm~\ref{alg:abduction} we can check that 
$B_\hbar=H$ and $E_\hbar=D$, since the only finding in the matching relation is 
$m^{Pw}$. Deduction then has to be performed after this last finding, using the 
production $D \rightarrow q_{QRS} E$. After constraint checking, the resulting 
finding is as follows:

\begin{equation*}
 m^q_{n+1} = m^{QRS} = \langle q_{QRS}, \varnothing, T^b_{QRS} \in
[0.400,0.520], T^e_{QRS} \in [0.450,0.660] \rangle
\end{equation*}

Figure~\ref{fig:full_interpretation} illustrates the outcome of this reasoning
process and the uncertainty in the temporal limits of the predicted finding,
which is now focused on in the interpretation $I_2$.
}
\end{example}

\subsection{Building an interpretation: Subsumption}

Subsumption is performed when the attention is focused on a finding previously 
deduced from some abstraction grammar (see Algorithm~\ref{alg:subsume}). This 
reasoning mode avoids the generation of a new hypothesis for every piece of 
available evidence if it can be explained by a previous hypothesis. The 
\begin{small}SUBSUME()\end{small} function explores the set of observations 
$\mathcal{O}$ and selects those consistent with the constraints on the finding 
in the focus of attention (line~\ref{ln:match}), expanding the matching relation 
of the corresponding hypothesis in different descendant interpretations 
(line~\ref{ln:matchexpand}). The focus of attention is then restored to its 
previous state (line~\ref{ln:frestore}), allowing the deduction of new findings 
from the same hypothesis. The \begin{small}SUBSUME()\end{small} function clearly 
enforces the simplicity principle. 

\begin{algorithm}
\captionsetup{font=small}
\caption{Moving forward an interpretation through subsumption.}
\label{alg:subsume}
\begin{algorithmic} [1]
\Function{subsume}{$I, m_i$}
    \State \textbf{var} $desc = \varnothing$
    \ForAll{$o_j \in \mathcal{O} ~|~ m_i \leftarrowtail o_j $} \label{ln:match}
      \State $I' = \langle O_I , P_I,  \leftarrowtail_I \cup ~\{m_i 
\leftarrowtail o_j\} \rangle$ \label{ln:matchexpand}
      \State \Call{get\_focus}{$I'$}.\Call{pop}{$m_i$} \label{ln:frestore}
      \State $desc = desc \cup \{I'\}$
    \EndFor
    \State \Return $desc$
  \EndFunction
\end{algorithmic}
\end{algorithm}

\begin{example}
{\em
Let us consider the interpretation $I_2$ shown in 
Figure~\ref{fig:full_interpretation}. If we apply the subsumption procedure, it 
is possible to set a matching relation between $o^{QRS}$ and $m^{QRS}$, since 
this observation satisfies all the constraints on the finding. The result is 
shown in the interpretation $I_3$. Note that the uncertainty in the end time of 
the $o^N$ hypothesis is now reduced after the matching, having $T^e_N \in 
[0.631, 1.030]$. Following this, the attention focuses once again on this 
hypothesis, and a new deduction operation may be performed.
}
\end{example}

\subsection{Building an interpretation: Prediction}
\label{sec:deduce_m}
This reasoning mode is also performed when the attention is focused on a finding 
deduced from some abstraction grammar (see Algorithm~\ref{alg:prediction}). In 
this case, if a finding previously deduced has not yet been observed, it will be 
predicted.

\begin{algorithm}
\captionsetup{font=small}
\caption{Moving forward an interpretation through the prediction of 
non-available evidence.}
\label{alg:prediction}
\begin{algorithmic} [1]
\Function{predict}{$I, m_i$}
    \State \textbf{var} $desc = \varnothing$
    \ForAll{$G_h = \langle V_N, V_T, H, R \rangle \in \mathcal{G} ~|~ h 
~\mathtt{is\_a}~ q(m_i)$} \label{ln:gselect}
      \State $P_H = \langle h, M_H=\varnothing, C_H=\varnothing, 
\varTheta_H=\varnothing \rangle$ \label{ln:emptypattern}
      \State $\hbar = \langle o_h, P_H, \leftarrowtail_\hbar=\varnothing 
\rangle$ \label{ln:emptymatch}
      \State $L_\hbar = \varnothing; B_\hbar=E_\hbar=H$ \label{ln:BEH}
      \State $I' = \langle O_I \cup \{o_h\}, P_I \cup \{P_H\},
\leftarrowtail_I \cup ~\{m_i \leftarrowtail o_h\} \rangle$
      \State $\mathcal{O} = \mathcal{O} \cup \{o_h\}$
      \State \Call{get\_focus}{$I'$}.\Call{pop}{$m_i$} \label{ln:dpop}
      \State \Call{get\_focus}{$I'$}.\Call{push}{$o_h$} \label{ln:dpush}
      \State $desc = desc \cup \{I'\}$
    \EndFor    
    \State \Return $desc$
  \EndFunction
\end{algorithmic}
\end{algorithm}

The goal of the \begin{small}PREDICT()\end{small} function is to conjecture a 
new observation to match the focused finding. For this, the abstraction model is 
explored and those grammars whose hypothesized observable is more specific than 
the predicted observable are selected (line~\ref{ln:gselect}). Then, a new 
pattern is initialized with no evidence supporting it, and a new abstraction 
hypothesis with an empty matching relation is generated 
(lines~\ref{ln:emptypattern}-\ref{ln:emptymatch}). Finally, the attention 
focuses on the observation being guessed (lines~\ref{ln:dpop}-\ref{ln:dpush}) to 
enable the \begin{small}DEDUCE()\end{small} function to start a new test step at 
a lower abstraction level. Since $L_\hbar$ is initialized as an empty list 
(line~\ref{ln:BEH}), $B_\hbar$ and $E_\hbar$ point to the initial symbol of the 
grammar, and the corresponding abstraction pattern will be generated only 
towards the end of the grammar.

\begin{example}
{\em
 Starting from the $I_3$ interpretation shown in 
Figure~\ref{fig:full_interpretation}, the next step we can take to move forward 
the interpretation is a new deduction on the $o^N$ hypothesis, generating a new 
finding $m^{Tw}$ and leading to the $I_4$ interpretation. Since there is no 
available observation of the T wave, a matching with this new finding $m^{Tw}$ 
cannot be made by the \begin{small}SUBSUME()\end{small} function, thus, the only 
option for moving forward this interpretation is through prediction. Following 
the \begin{small}PREDICT()\end{small} function, the $G_{Tw}$ grammar can be 
selected, and a new observation $o^{Tw}$ can be conjectured, generating the 
$I_5$ interpretation.

From $I_5$ we can continue the deduction on the $o^{Tw}$ hypothesis. If we apply 
the \begin{small}DEDUCE()\end{small} function we obtain the $m^{QRS'}$ finding 
from the environment, shown in Figure~\ref{fig:full_interpretation} as $I_6$.  
To move on, we can apply the \begin{small}SUBSUME()\end{small} function, 
establishing the matching relation $\{m^{QRS'} \leftarrowtail o^{QRS}\}$. This 
leads to the $I_7$ interpretation, in which the uncertainty on the $o^{Tw}$ 
observation is reduced; however, the evidence for the $P_{Tw}$ pattern is not 
yet complete. A new \begin{small}DEDUCE()\end{small} step is necessary, which 
deduces the $m^{wave}$ necessary finding in the $I_8$ interpretation. This 
finding is also absent, so another \begin{small}PREDICT()\end{small} step is 
required. In this last step, the $P_{wave}$ pattern can be applied to observe 
the deviation in the raw ECG signal, generating the $o^{wave}_3$ observation and 
completing the necessary evidence for the $o^{Tw}$ observation and thus also for 
$o^N$. Constraint solving assigns the value of $t^b_{Tw}$, $t^e_{Tw}$ and 
$t^e_N$, so the result is a cover of the initial interpretation problem in which 
all the hypotheses have a necessary and sufficient set of evidence. This 
solution is depicted in $I_9$. 

It is worth noting that in this example the global matching relation 
$\leftarrowtail_I$ is not injective, since $m^{QRS} \leftarrowtail o^{QRS}$ and 
$m^{QRS'} \leftarrowtail o^{QRS}$. Also note that each interpretation only 
generates one descendant; in a more complex scenario, however, the possibilities 
are numerous, and the responsibility of finding the proper sequence of reasoning 
steps lies with the \begin{small}CONSTRUE()\end{small} algorithm.
}

\begin{figure}[ht!]
  \centering
  \begin{subfigure}[b]{0.45\textwidth}
    \includegraphics[width=\textwidth]{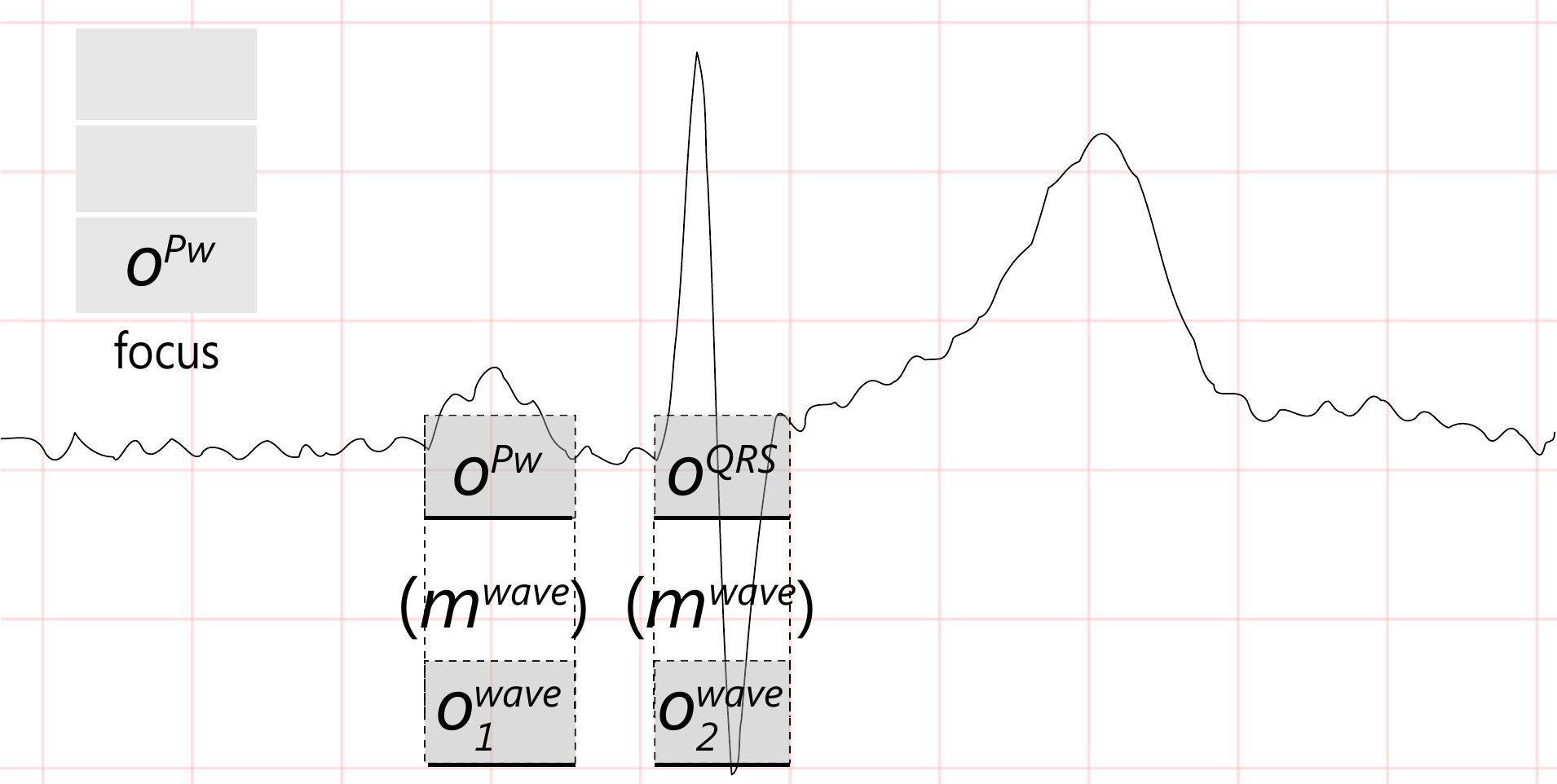}
    \caption*{$I_0$ = Initial evidence}
  \end{subfigure}
  \quad 
  \begin{subfigure}[b]{0.45\textwidth}
      \includegraphics[width=\textwidth]{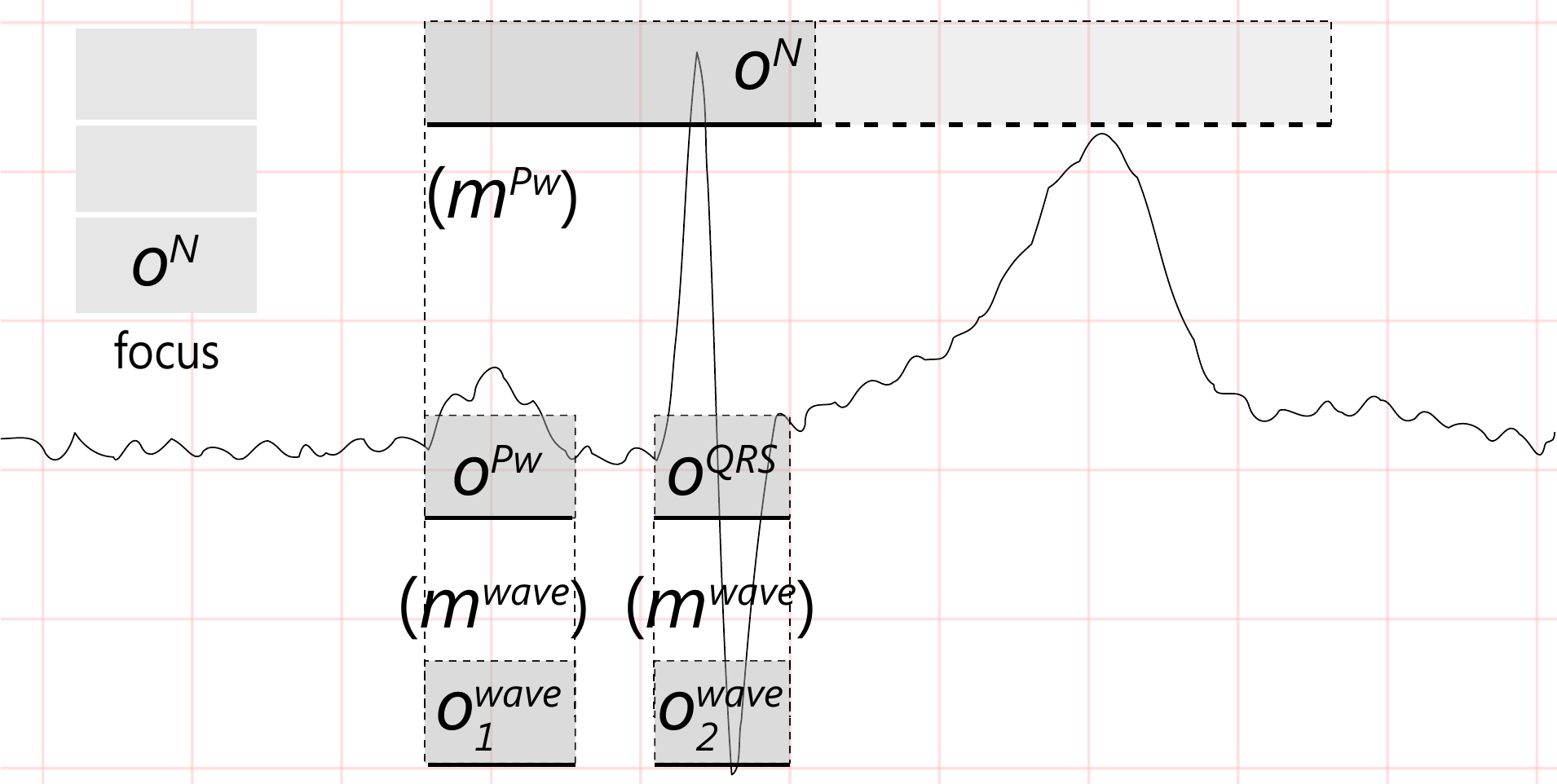}
      \caption*{$I_1$ = \scriptsize{ABDUCE($I_0, o^{Pw}$)}}
  \end{subfigure}
  \vspace{0.7em}
  
  \begin{subfigure}[b]{0.45\textwidth}
    \includegraphics[width=\textwidth]{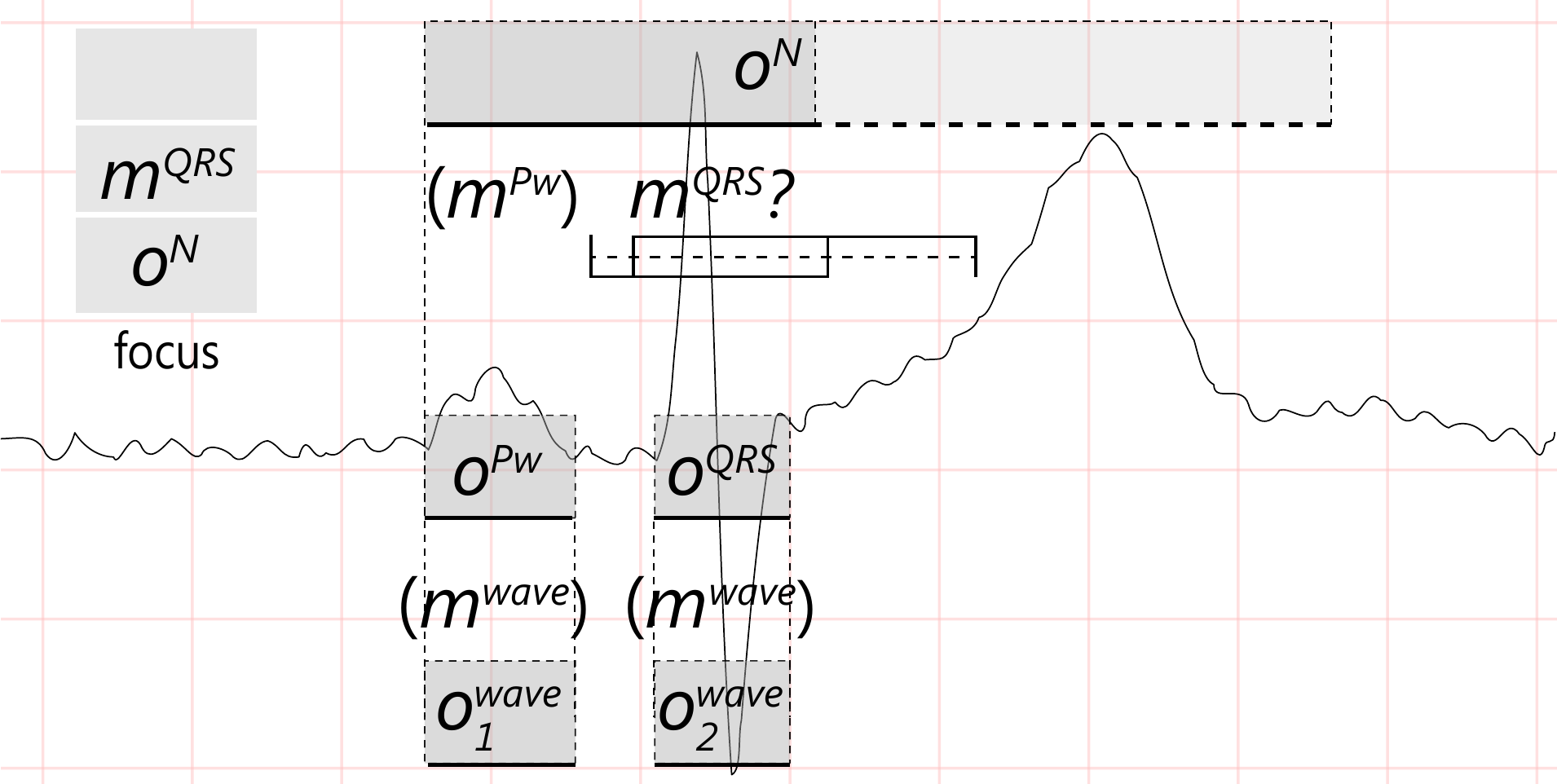}
    \caption*{$I_2$ = \scriptsize{DEDUCE($I_1, o^N$)}}
  \end{subfigure}
  \quad 
  \begin{subfigure}[b]{0.45\textwidth}
      \includegraphics[width=\textwidth]{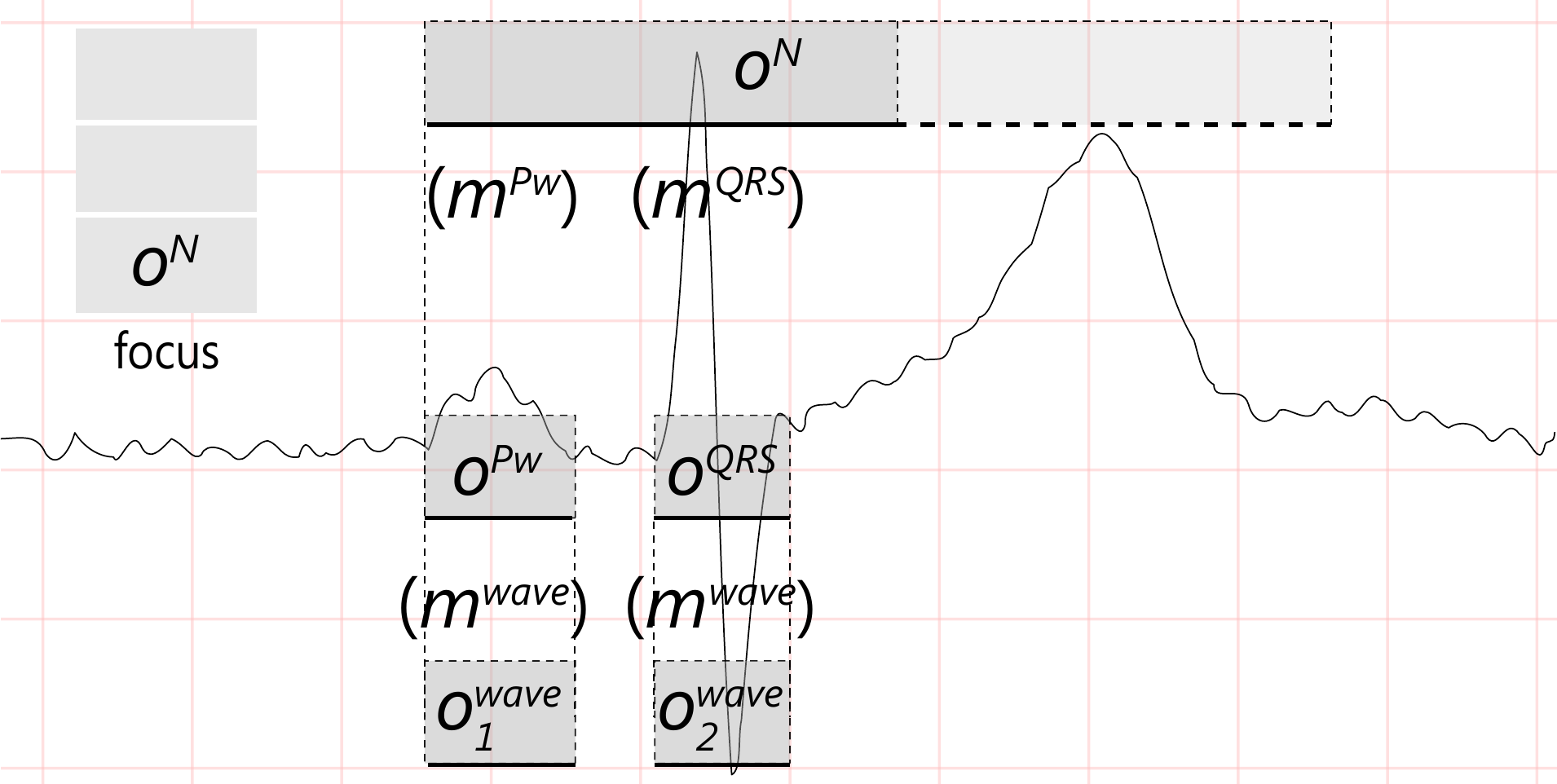}
      \caption*{$I_3$ = \scriptsize{SUBSUME($I_2, m^{QRS}$)}}
  \end{subfigure}
  \vspace{0.7em}
  
  \begin{subfigure}[b]{0.45\textwidth}
    \includegraphics[width=\textwidth]{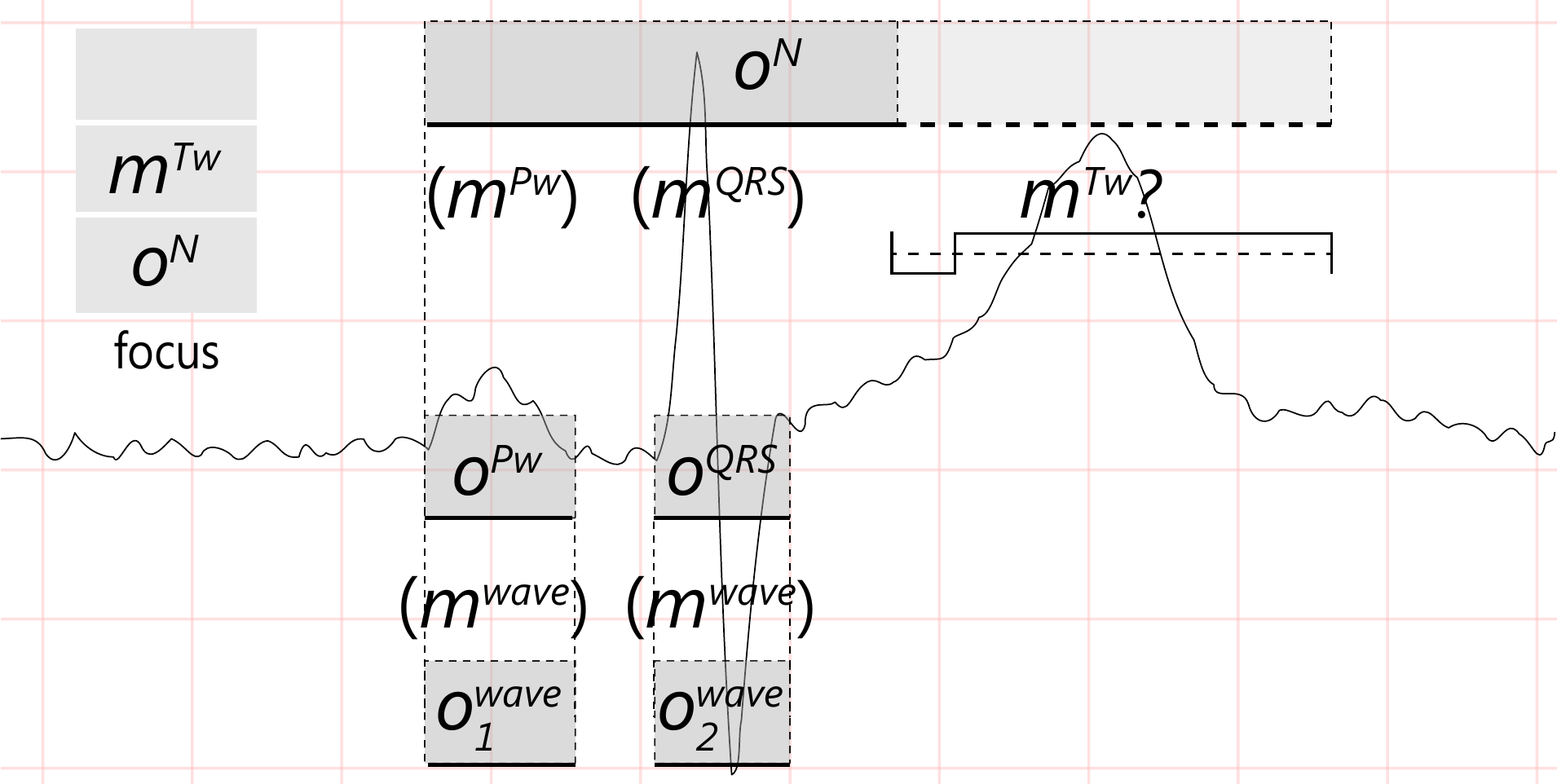}
    \caption*{$I_4$ = \scriptsize{DEDUCE($I_3, o^N$)}}
  \end{subfigure}
  \quad 
  \begin{subfigure}[b]{0.45\textwidth}
      \includegraphics[width=\textwidth]{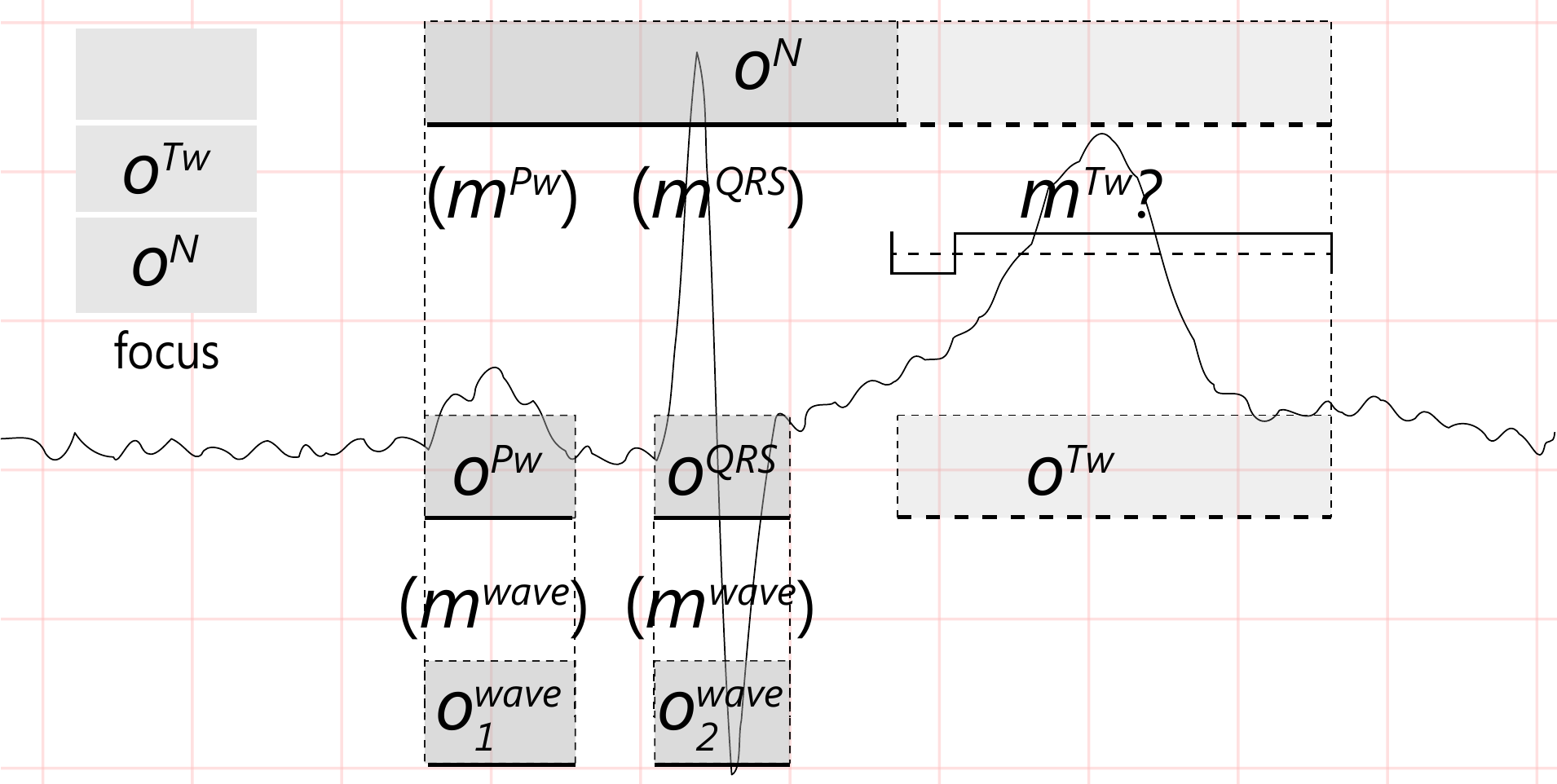}
      \caption*{$I_5$ = \scriptsize{PREDICT($I_4, m^{Tw}$)}}
  \end{subfigure}
  \vspace{0.7em}
  
  \begin{subfigure}[b]{0.45\textwidth}
    \includegraphics[width=\textwidth]{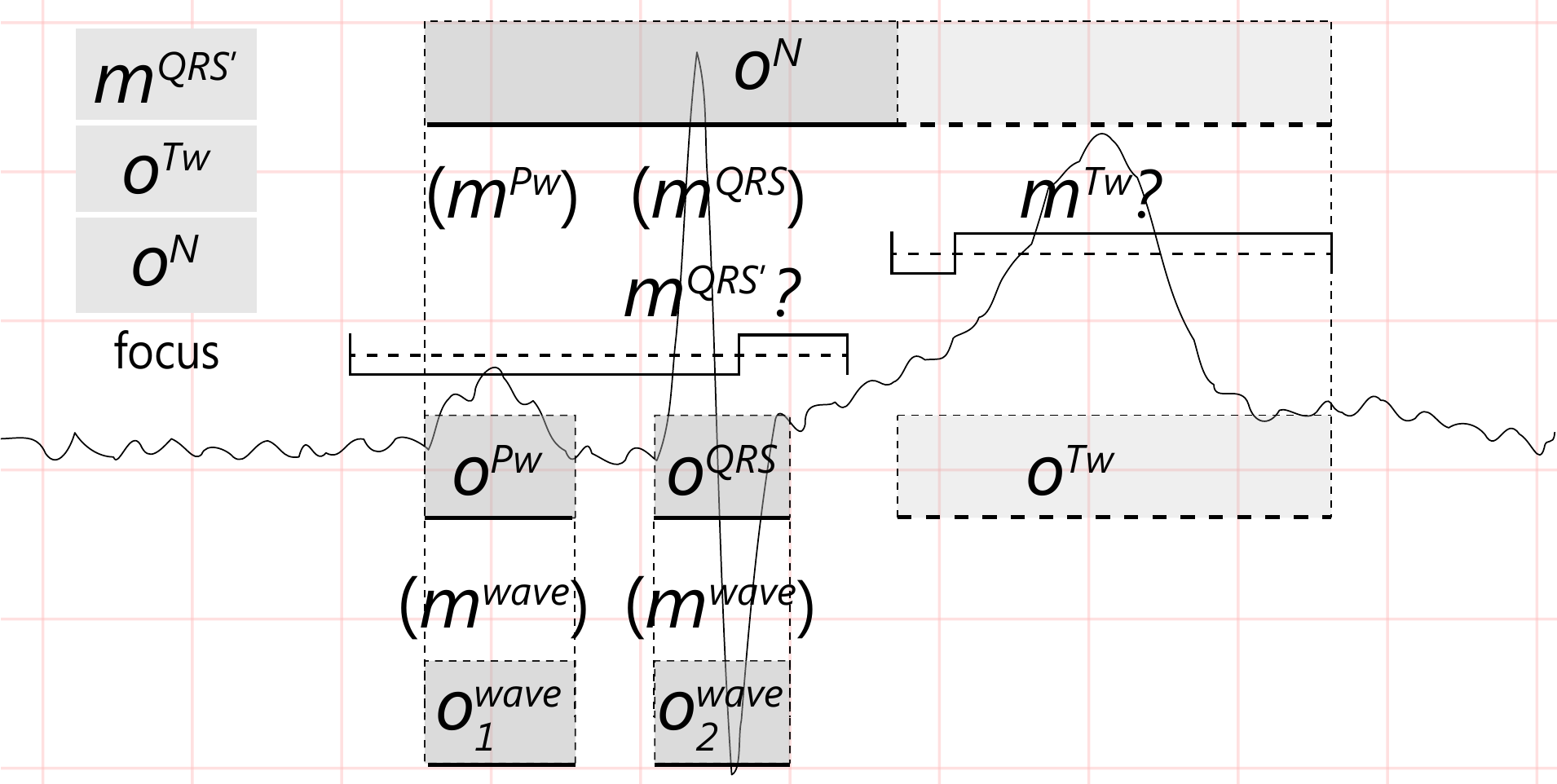}
    \caption*{$I_6$ = \scriptsize{DEDUCE($I_5, o^{Tw}$)}}
  \end{subfigure}
  \quad 
  \begin{subfigure}[b]{0.45\textwidth}
      \includegraphics[width=\textwidth]{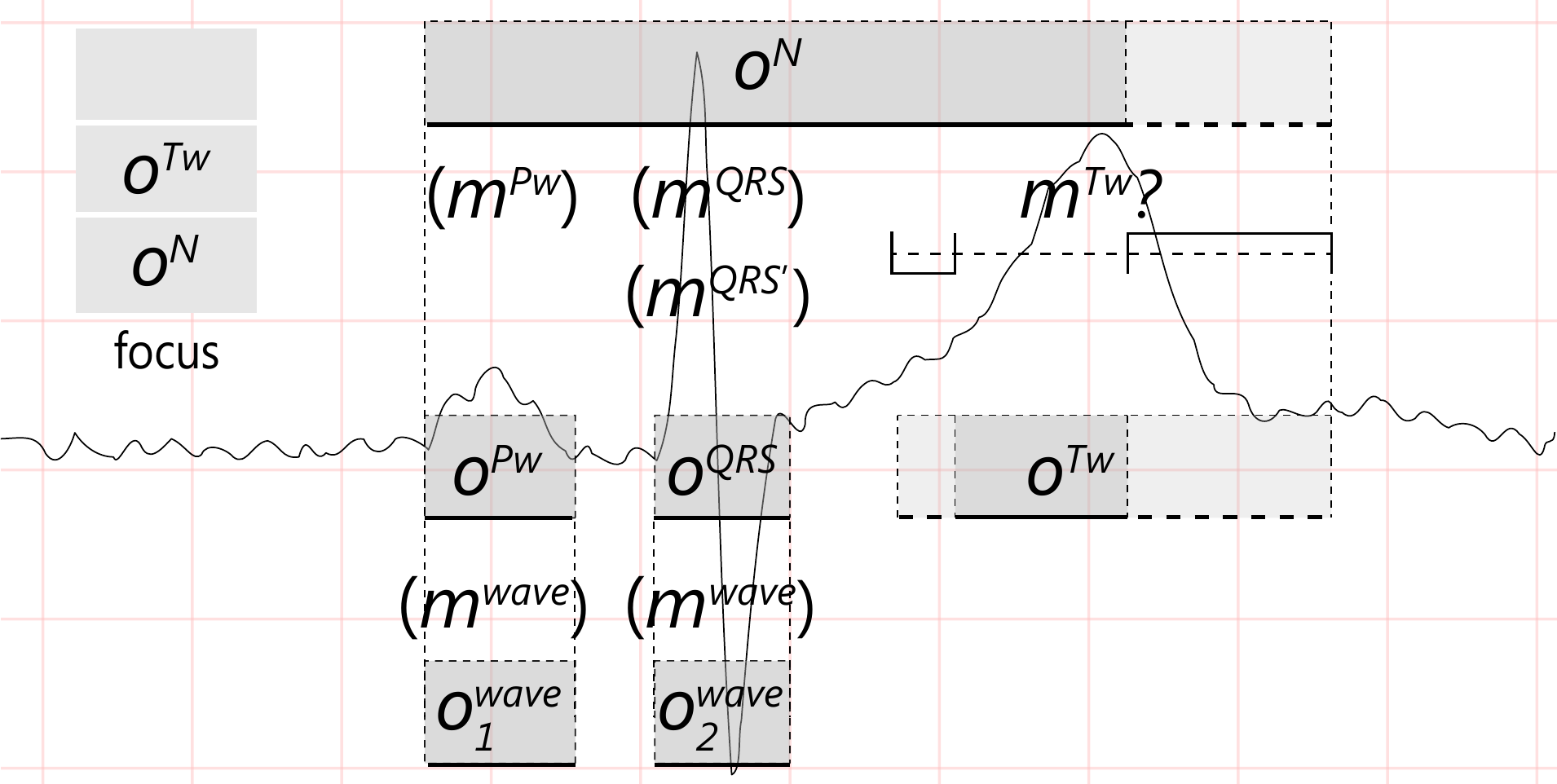}
      \caption*{$I_7$ = \scriptsize{SUBSUME($I_6, m^{QRS'}$)}}
  \end{subfigure}
  \vspace{0.7em}
  
  \begin{subfigure}[b]{0.45\textwidth}
    \includegraphics[width=\textwidth]{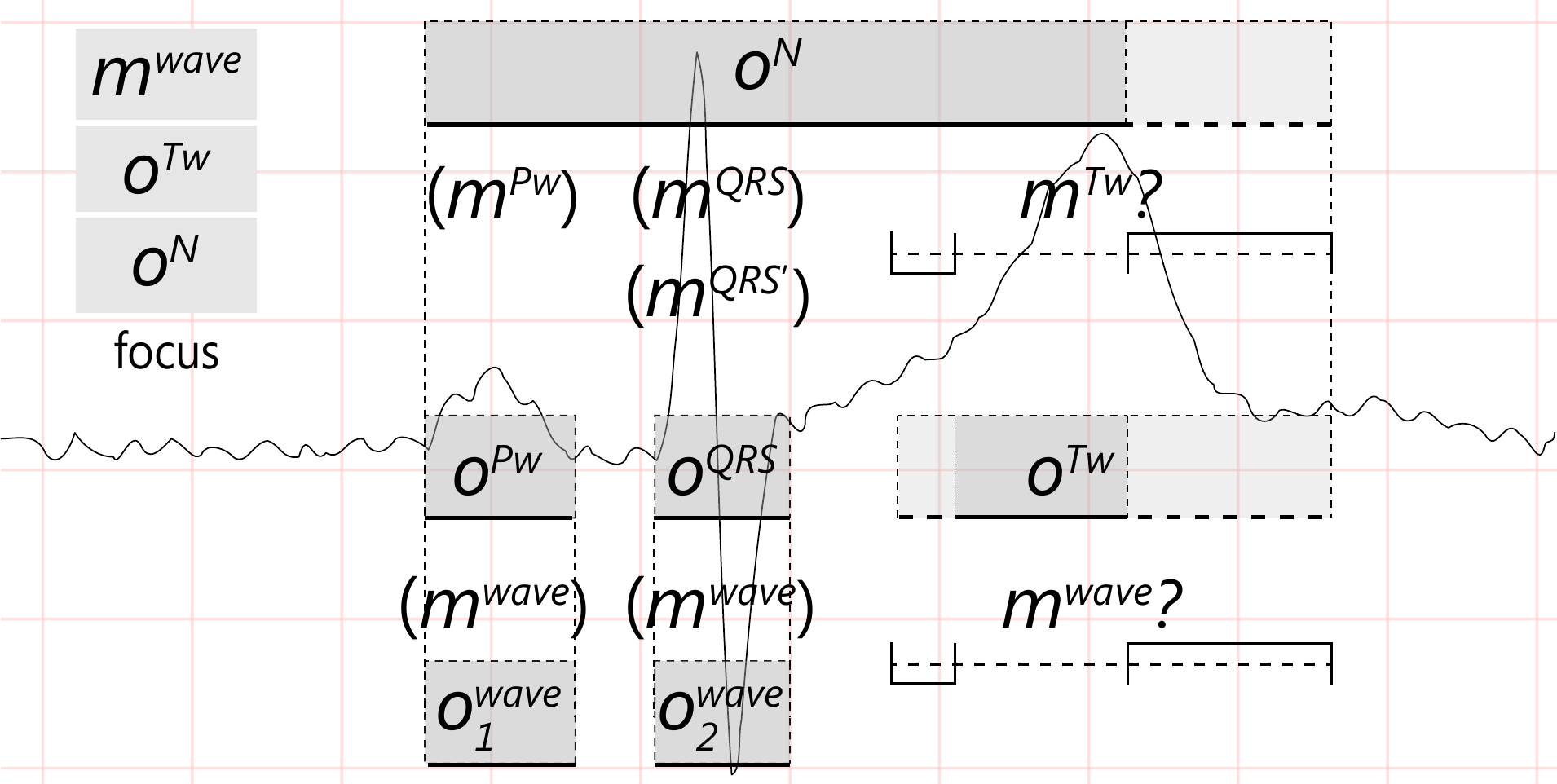}
    \caption*{$I_8$ = \scriptsize{DEDUCE($I_7, o^{Tw}$)}}
  \end{subfigure}
  \quad 
  \begin{subfigure}[b]{0.45\textwidth}
      \includegraphics[width=\textwidth]{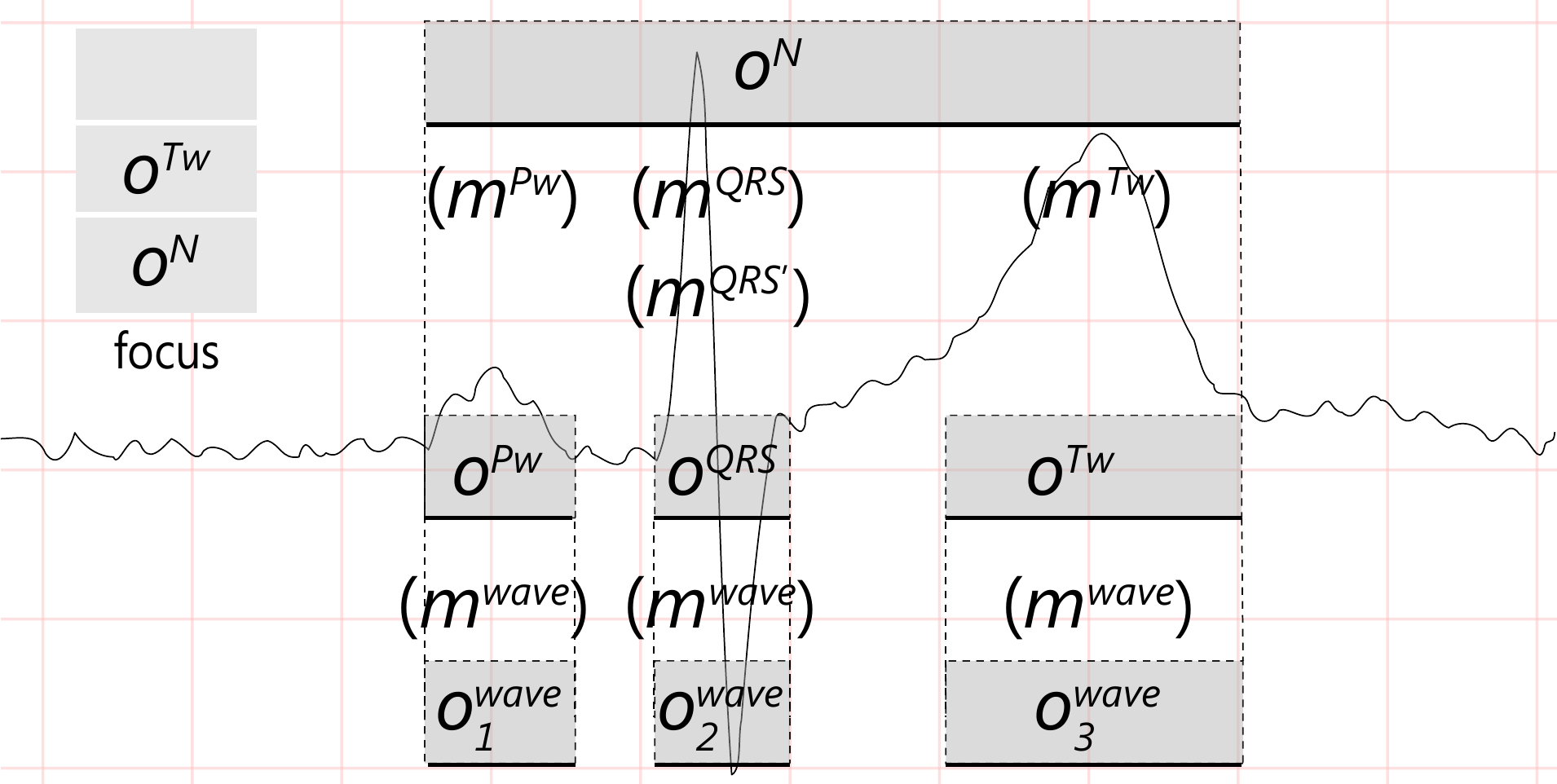}
      \caption*{$I_9$ = \scriptsize{PREDICT($I_8, m^{wave}$)}}
  \end{subfigure}
  
\caption{Sequence of reasoning steps for solving a simple interpretation 
problem.} 
\label{fig:full_interpretation}
\end{figure}
\end{example}

\subsection{Improving the efficiency of interpretation through saliency}
Starting a hypothesize-and-test cycle for every single sample is not feasible 
for most of the time series interpretation problems. Still, many problems may 
benefit from certain saliency features that can guide the attention focus to 
some limited temporal fragments that can be easily interpretable. Thus, the 
interpretation of the whole time series can pivot on a reduced number of initial 
observations, thereby speeding up the interpretation process.

A saliency-based attentional strategy can be devised from the definition of 
abstraction patterns using a subset of their constraints as a coarse filter to 
identify a set of plausible observations. For example, in the ECG interpretation 
problem the most common strategy is to begin the analysis by considering a 
reduced set of time points showing a significant slope in the signal, consistent 
with the presence of QRS complexes~\cite{Zong03}. This small set of evidence 
allows us to focus the interpretation on the promising signal segments, in the 
same way that a cardiologist focuses on the prominent peaks to start the 
analysis~\cite{Marriott08}. It should be noted that this strategy is primarily 
concerned with the behavior of the focus of attention, and that it does not 
discard the remaining, non-salient observations, as these are included later in 
the interpretation by means of the subsumption and prediction reasoning modes.

\section{Advantages of the framework}
\label{sec:strengths}

In this section we provide several practical examples which illustrate some of 
the strengths of the proposed interpretation framework and its ability to 
overcome typical weaknesses of the strategies based solely on a classification 
approach.

\subsection{Avoiding a casuistry-based interpretation}

In the time domain, classification-based recognition of multiple processes 
occurring concurrently usually leads to a casuistry-based proliferation of 
classes, in which a new class is usually needed for each possible superposition 
of processes in order to properly identify all situations. It is common to use a 
representation in the transform domain, where certain regular processes are 
easily separable, although at the expense of a cumbersome representation of the 
temporal information~\cite{Morchen03}. In contrast, in the proposed framework, 
the hypothesize-and-test cycle aims to conjecture those hypotheses that best 
explain the available evidence, including simultaneous hypotheses in a natural 
way as long as these are not mutually exclusive.

\begin{figure}[h!]  
 \centering
  \includegraphics[width=\textwidth]{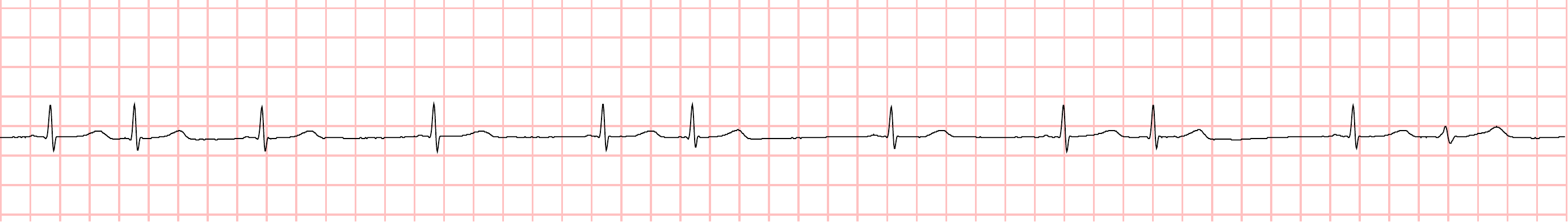}
 \caption{False atrial fibrillation episode. [Source: Mobiguide 
Project~\cite{Sacchi15}, private recording]}
 \label{fig:false_afib}  
\end{figure}

ECG interpretation provides some interesting examples of this type of problem. 
Atrial fibrillation, a common heart arrhythmia caused by the independent and 
erratic contractions of the atrial muscle fibers, is characterized by an 
irregularly irregular heart rhythm~\cite{Marriott08}. Most of the classification 
techniques for the identification of atrial fibrillation are based on the 
analysis of the time interval between consecutive beats, and attempt to detect 
this irregularity~\cite{Petrenas15}. These techniques offer good results in 
those situations in which atrial fibrillation is the only anomaly, but they fail 
to properly identify complex scenarios which go beyond the distinction between 
atrial fibrillation and normal rhythm. In the strip shown in  
Figure~\ref{fig:false_afib}, obtained during a pilot study for the home 
follow-up of patients with cardiac diseases~\cite{Sacchi15}, such a classifier 
would wrongly identify this segment as an atrial fibrillation episode, since the 
observed rhythm variability is consistent with the description of this 
arrhythmia. In contrast, the present interpretation framework correctly explains 
the first five beats as a sinus bradycardia, compatible with the presence of a 
premature ectopic beat in the second position, followed by a trigeminy pattern 
during six beats, and finally another ectopic beat with a morphology change. The 
reason to choose this interpretation, despite being more complex than the atrial 
fibrillation explanation, is that it is able to abstract some of the small P 
waves before the QRS complexes, increasing the interpretation coverage. 


\subsection{Coping with ignorance}

Most of the classifiers solve a separability problem among classes, either by 
learning from a training set or by eliciting prior knowledge, and these are 
implicitly based on the closed-world assumption, i.e., every new instance to be 
classified is assigned to one of the predefined classes. Such classifiers may 
additionally include a 'reject' option for all those instances that could be 
misclassified since they appear too close to the classification boundaries 
\cite{Chow70,Fumera00}. This reject option is added as another possible answer 
expressing doubt. However, such classifiers fail to classify new instances of 
unknown classes, since they cannot express ignorance. An approach to this 
problem can be found in novelty detection proposals \cite{Pimentel14}, which 
can detect when a new instance does not fit any of the predefined classes 
as it substantially differs from those instances available during training. 
Still, these are limited to a common feature representation for every 
instance, hindering the identification of what is unintelligible from the 
available knowledge.

The proposed framework provides an expression of ignorance as a common result of 
the interpretation problem. As long as the abstraction model is incomplete, the 
non-coverage of some piece of evidence by any interpretation is an expression of 
partial ignorance. In the extreme case, the trivial interpretation $I_0$ may be 
a correct solution of an interpretation problem, expressing total ignorance. 
Furthermore, abduction naturally includes the notion of ignorance in the 
reasoning process, since any single piece of evidence can be sufficient to guess 
an interpretation, and the hypothesize-and-test cycle can be understood as a 
process of incremental addition of evidence against an initial state of 
ignorance, while being able to provide an interpretation at any time based on 
the available evidence.

\begin{figure}[h!]  
 \centering
  \includegraphics[width=\textwidth]{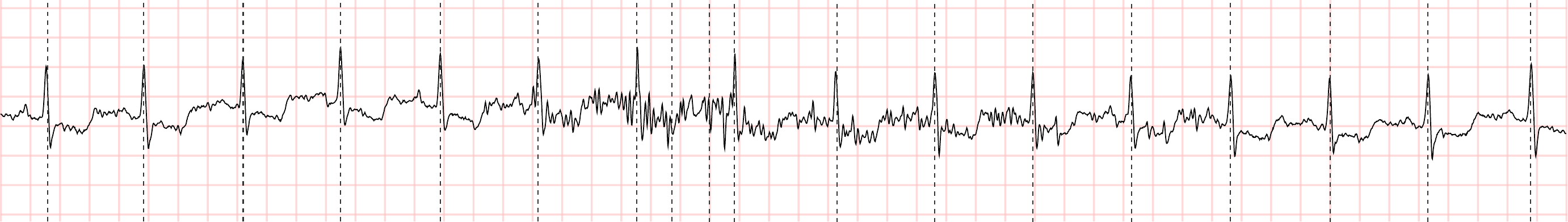}
 \caption{Unintelligible evidence due to noise. [Source: MIT-BIH arrhythmia DB, 
recording: 112, between 13:46.200 and 13:56.700]}
 \label{fig:noisy_112}  
\end{figure}

As an example, consider the interpretation problem illustrated in 
Figure~\ref{fig:noisy_112}. Let the initial evidence be the set of QRS 
annotations obtained by a state-of-the art detection algorithm~\cite{Zong03}. In 
this short interval, the eighth and ninth annotations correspond to false 
positives caused by noise. A classification-based strategy processes these two 
annotations as true QRS complexes, and the monotone nature of the reasoning 
prevents their possible refutation, probably leading to beat misclassification 
and false arrhythmia detection, with errors propagating onwards to the end of 
the processing. In contrast, the present framework provides a single normal 
rhythm as the best interpretation, which explains all but the two aforementioned 
annotations, which are ignored and considered unintelligible in the available 
model. It is also worth noting the ability of this framework to integrate the 
results of an available classifier as a type of constraint specification in the 
interpretation cycle.

\subsection{Looking for missing evidence}

The application of the classification paradigm to pattern detection also entails 
the potential risk of providing false negative results. In the worst case, a 
false negative result may be interpreted by a decision maker as evidence of 
absence, leading to interpretation errors with their subsequent costs, or in the 
best case as an absence of evidence caused by the lack of a proper detection 
instrument.

Even though abduction is fallible, and false negative results persist, the 
hypothesize-and-test cycle involves a prediction mechanism that points to
missing evidence that is expected and, moreover, estimates when it should 
appear. Both the bottom-up and top-down processing performed in this 
cycle reinforces confidence in the interpretation, since the semantics of 
any conclusion is widened according to its explanatory power.

\begin{figure}[h!]  
 \centering
  \includegraphics[width=\textwidth]{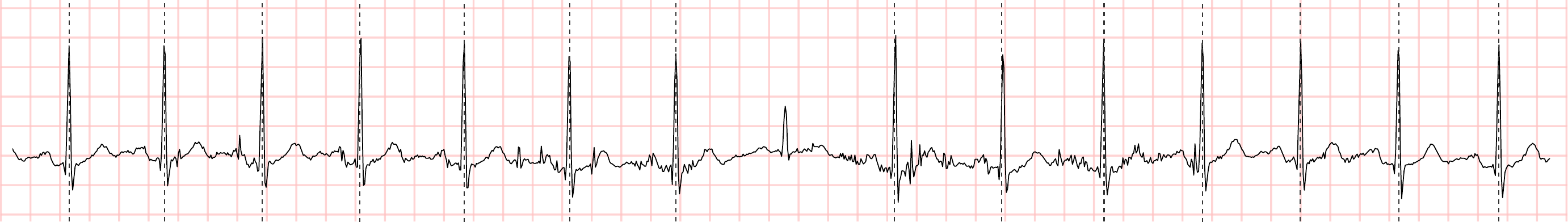}
 \caption{Missing evidence that may be discovered by prediction.  [Source: 
MIT-BIH normal sinus rhythm DB, recording: 18184, between 09:12:45.000 and 
09:12:55.500]}
 \label{fig:missb_18184}  
\end{figure}

As an example, consider the interpretation problem illustrated in Figure 
\ref{fig:missb_18184}. The initial evidence is again a set of QRS annotations 
obtained by a state-of-the-art detection algorithm~\cite{Zong03}. Note 
that the eighth beat has not been annotated, due to a sudden decrease in the 
signal amplitude. This error can be amended in the hypothesize-and-test cycle, 
since the normal rhythm hypothesis that abstracts the first seven QRS 
annotations predicts the following QRS to be in the position of the missing 
annotation, and the \begin{small}PREDICT()\end{small} procedure can look for 
this (e.g., checking an alternative set of constraints).

The capability of abduction to ignore or look for new evidence has been tested 
with a simplified version of the present framework in the QRS detection problem 
~\cite{Teijeiro14}, leading to a statistically significant improvement over a 
state-of-the art algorithm.

\subsection{Interpretability of the reasoning process and the results}

The interpretability of a reasoning formalism, defined as the ability to 
understand and evaluate its conclusions, is an essential feature for achieving 
an adequate confidence in decision making \cite{Nauck99}. In this sense, there 
are a number of classification methods with good interpretability; however, the 
methods that typically offer the best performance belong to the so-called 
\textit{black box} approaches.

The present interpretation framework is able to provide a justification of any 
result in relation to the available model. Given any solution or partial 
solution of an interpretation problem, the searching path up to $I_0$ gives 
full details of all the reasoning steps taken to this end, and any abstraction 
hypothesis can be traced back to the information supporting it.

This interpretation framework is also able to answer the question of why a 
certain hypothesis has been rejected or neglected at any reasoning step. This is 
done by exploring the branches outside the path between $I_0$ and the solution. 
Since the $K$ exploration parameter within the 
\begin{small}CONSTRUE()\end{small} algorithm has been chosen as the maximum 
number of hypotheses that may explain a given observable, it is possible to 
reproduce the reasoning steps taken in the conjecture of any abstraction 
hypothesis, and to check why this did not succeed (non-satisfaction of pattern 
constraints, lower coverage, etc.). This can be useful in building and refining 
the knowledge base.

\section{Experimental evaluation: beat labeling and arrhythmia detection}
\label{sec:evaluation}

The interpretation of electrocardiograms has served both as a challenge and as 
an inspiration for the AI community due to a number of factors that can be 
summarized as: (1) the complexity of the physiological processes underlying what 
is observed; and (2) the absence of an accurate model of the heart and the 
hardly formalizable knowledge that constitutes the experience of the 
cardiologist. There are numerous problems falling within the scope of ECG 
interpretation, the most relevant being heartbeat labeling~\cite{Luz15}. We have 
tested the present framework by abductively identifying and measuring a set of 
qualitative morphological and rhythm attributes for each heartbeat, and using a 
rule-based classifier to assign a label to clusters of similar 
heartbeats~\cite{Teijeiro16}. It is noteworthy that an explicit representation 
of knowledge has been adopted, namely the kind of knowledge that can be found in 
an ECG handbook. Table~\ref{tab:comparison} reproduces the performance 
comparison between this approach and the most relevant automatic and assisted 
approaches of the state-of-the art, using sensitivity and positive predictivity 
of ventricular and supraventricular ectopic beat classes.

\begin{table}
\renewcommand{\arraystretch}{1.3}
\setlength{\tabcolsep}{3.5pt}
\caption{VEB and SVEB classification performance of the abductive approach and 
comparison with the most relevant automatic and assisted methods of the 
state-of-the-art}
\label{tab:comparison}
\centering
\scriptsize
\begin{tabular}{c c r r r r}
\hline \hline
 & & \multicolumn{2}{c}{VEB} & \multicolumn{2}{c}{SVEB}\\
\cline{3-6}
Dataset & Method & $Se$ & $P^+$ & $Se$ & $P^+$\\
\hline
\multirow{5}{80pt}{\centering{MIT-BIH Arrhythmia DS1+DS2}} & Teijeiro 
\textit{et al.} - Automatic & 92.82 & 92.23 & 85.10 & 84.51\\
 & Llamedo \textit{et al.} - Assisted & 90$\pm$1 & 97$\pm$0 & 89$\pm$2 & 
88$\pm$3\\
 & Kiranyaz \textit{et al.} - Assisted & 93.9 & 90.6 & 60.3 & 63.5\\
 & Ince \textit{et al.} - Assisted & 84.6 & 87.4 & 63.5 & 53.7\\
 & Llamedo \textit{et al.} - Automatic & 80$\pm$2 & 82$\pm$3 & 76$\pm$2 & 
43$\pm$2\\
\hline
\multirow{7}{80pt}{\centering{MIT-BIH Arrhythmia DS2}} & Teijeiro \textit{et 
al.} - Automatic & 94.63 & 96.79 & 87.17 & 83.98\\
 & Llamedo \textit{et al.} - Assisted & 93$\pm$1 & 97$\pm$1 & 92$\pm$1 & 
90$\pm$3\\
 & Kiranyaz \textit{et al.} - Assisted & 95.0 & 89.5 & 64.6 & 62.1\\
 & Chazal \textit{et al.} - Assisted & 93.4 & 97.0 & 94.0 & 62.5\\
 & Zhang \textit{et al.} - Automatic & 85.48 & 92.75 & 79.06 & 35.98\\
 & Llamedo \textit{et al.} - Automatic & 89$\pm$1 & 87$\pm$1 & 79$\pm$2 & 
46$\pm$2\\
 & Chazal \textit{et al.} - Automatic & 77.7 & 81.9 & 75.9 & 38.5\\
\hline \hline
\end{tabular}
\end{table}

As it can be seen, this method significantly outperforms any other automatic 
approaches in the state-of-the-art, and even improves most of the assisted 
approaches that require expert aid. The most remarkable improvement concerns the 
classification of supraventricular ectopic beats, which are usually hard to 
distinguish using only morphological features. The abductive interpretation in 
multiple abstraction levels, including a rhythm description of signal, is what 
enables a more precise classification of each individual heartbeat.

Furthermore, the abductive interpretation approach has been used for arrhythmia 
detection in short single-lead ECG records, focusing on atrial 
fibrillation~\cite{Teijeiro17}. The interpretation results are combined with 
machine learning techniques to obtain an arrhythmia classifier, achieving the 
best score in the 2017 Physionet/CinC Challenge dataset and outperforming some 
of the most popular techniques such as deep learning and random 
forests~\cite{Clifford17}.

\section{Discussion}
\label{sec:discussion}

A new model-based framework for time series interpretation is proposed. This 
framework relies on some basic assumptions: (i) interpretation of the behavior 
of a system from the set of available observations is a sort of conjecturing, 
and as such follows the logic of abduction; (ii) the interpretation task 
involves both bottom-up and top-down processing of information along a set of 
abstraction levels; (iii) at the lower levels of abstraction, the interpretation 
task is a form of precompiled knowledge-based pattern recognition; (iv) the 
interpretation task involves both the representation of time and reasoning about 
time and along time.

Model-based representation in the present framework is based on the notion of 
abstraction pattern, which defines an abstraction relation between observables 
and provides the knowledge and methods to conjecture new observations from 
previous ones. Let us deepen in both the backward and forward logical meaning of 
an abstraction pattern, following a reasoning similar to that of 
\cite{Brusoni98}:

\begin{itemize}
 \item {\em Backward meaning}. From the backward reading of an abstraction 
pattern $P$, a hypothesis $h$ is a possible abstraction of $m_1,\ldots,m_n$, 
provided that the constraints in $C_P$ hold. An abstraction pattern satisfies 
the {\em compositionality principle} of abductive reasoning, and hence an 
abstraction hypothesis can be conjectured from a single piece of evidence, and 
new pieces of evidence can be added later \cite{Flach96}. On the other hand, if 
there are multiple ways of observing $h$ by means of multiple patterns, and 
their respective constraints are inconsistent with the evidence, we do not 
conclude $\neg h$, interpreted as failure to prove $h$; we will only conclude 
$\neg h$ in all those interpretations conjecturing an observation of a different 
$h'$, where $h$ and $h'$ are mutually exclusive.

 \item {\em Forward meaning}. An abductive observation is built upon an 
archetypical representation of a hypothesis $h$, creating an observation as an 
instance of $h$ by estimating, from the available evidence, its attribute values 
$\mathbf{A}$ and its temporal location $T^b$ and $T^e$ by means of an 
observation procedure $\varTheta_P$. From a forward reading, assuming $h$ is 
true, there is an observation for each observable of the set $m_1,\ldots,m_n$ 
such that the constraints in $C_P$ hold. However, the estimated nature of 
abstraction does not usually allow us to infer, from the observation of $h$, the 
same observations of $m_1,\ldots,m_n$ that have been abstracted into $h$. We 
must presume instead that assuming $h$ is true entails the occurrence of an 
observation for each observable of $m_1,\ldots,m_n$, without necessarily 
entailing its attribute values and its temporal location.
\end{itemize}

Both the forward and the backward meanings of an abstraction pattern support the 
incremental building of an interpretation in the present framework. Thus, what 
initially was defined as a set covering problem of a time series fragment -a 
completely intractable problem as it moves away from a toy example- can be 
feasibly solved if it is properly structured in a set of abstraction levels, on 
which four reasoning modes (abduction, deduction, subsumption and prediction) 
can make a more efficient search of the best explanation under a parsimony 
criterion. Moreover, this incremental reasoning primarily follows the time 
direction, since the available knowledge is usually compiled in the form of a 
set of processes that can be expected to be found in a certain sequence, which 
underscores the anticipatory information contained in the evidence. 

An abstraction model, built on a set of abstraction patterns, establishes a 
causal responsibility for the behavior observed in a complex system 
\cite{Josephson94}. This responsibility is expressed in the language of 
processes: a process is said to be observable if it is assumed that it causes a 
recognizable trace in the physical quantity to be interpreted. This notion of 
causality is behind perception, i.e., concerned with the explanation of sensory 
data, in contrast with the notion of causality in diagnosis, concerned with the 
explanation of abnormality \cite{Console91a}.

Representing and reasoning about context is a relevant issue in model-based 
diagnosis \cite{Brusoni98, Console91a,Peng90,Shahar98}. A contextual observation 
is nothing more than another observation that need not be explained by a 
diagnosis. In most of the bibliography, the distinction between these two roles 
must be defined beforehand. Several other works enable the same observation to 
play different roles in different causal patterns, thus providing some general 
operations for expressing common changes made by the context in a diagnostic 
pattern \cite{Juarez08,Palma06}. In the present interpretation framework, an 
observation can either be part of the evidence to be explained in a certain 
abstraction pattern, or can be part of the environment in another abstraction 
pattern. Both types of observation play a part in the hypothesize-and-test 
cycle, with the only difference that observations of the environment of an 
abstraction pattern are not expected to be abstracted by this pattern. Hence, 
observations of the environment are naturally included in the deduction, 
subsumption and prediction modes of reasoning.

An important limitation of the present framework is its knowledge-intensive 
nature, requiring a non-trivial elicitation of expert knowledge. It is worth 
exploring different possibilities for the inclusion of machine learning 
strategies, both for the adaption and the definition of the knowledge base. A 
first approach may address the automatic adjustment of the initial constraints 
among recurrent findings in abstraction grammars. In this manner, for example, 
temporal constraints between consecutive heartbeats in a normal rhythm 
abstraction grammar could be adapted to the characteristics of the subject whose 
ECG is being interpreted, allowing the identification of possible deviations 
from normality with greater sensitivity. On the other hand, the discovery of new 
abstraction patterns and abstraction grammars by data mining methods appears 
as a key challenge. In this regard, the \begin{small}CONSTRUE()\end{small} 
algorithm should be extended by designing an \begin{small}INDUCE()\end{small} 
procedure aimed at conjecturing new observables after an inductive process. To 
this end, new default assumptions should be made in order to define those 
grammar structures that should rule the inductive process. These grammar 
structures may lead to discovery new morphologies or rhythms not previously 
included in the knowledge base. 

The proposed framework formulates an interpretation problem as an abduction 
problem with constraints, targeted at finding a set of hypotheses covering all 
the observations while satisfying a set of constraints on their attribute and 
temporal values. Thus, consistency is the only criterion to evaluate the 
plausibility of a hypothesis, resulting in a true or false value, and any evoked 
hypothesis (no matter how unusual it is) for which inconsistent evidence cannot 
be found is considered as plausible and, consequently, it will be explored in 
the interpretation cycle. Even though this simple approach has provided 
remarkable results, it can be expected that the inclusion of a hypothesis 
evaluation scheme, typically based on probability~\cite{Peng90, Poole00} or 
possibility~\cite{Dubois95, Palma06} theories, will allow us to better 
discriminate between plausible and implausible hypotheses, leading to better 
explanations with fewer computational requirements.

The expressiveness of the present framework should also be enhanced to support 
the representation of the \textit{absence} of some piece of evidence, in the 
form of negation, so that $\neg q$ represents the absence of $q$. The 
\textit{exclusion relation} is a first approach to manage with the notion of 
absence in the hypothesize-and-test cycle, since the occurrence of a process is 
negated by the concurrent occurrence of any of the processes related to it by 
the exclusion relation. On the other hand, an \textit{inhibitory relation} can 
enable us to represent a certain process preventing another from occurring under 
some temporal constraints, providing a method to insert the prediction of the 
absence of some observable in the hypothesize-and-test cycle. Furthermore, other 
forms of interaction between processes, possibly modifying the respective 
initial patterns of evidence, should be modeled.

Further efforts should be made to improve the efficiency of the interpretation 
process. To this end, two main strategies are currently being explored. In the 
fist strategy, the model structure is exploited to identify necessary and 
sufficient conditions for every hypothesis to be conjectured; the necessary 
conditions avoid the expansion of the hypotheses that can be ruled out because 
they are inconsistent with observations, while sufficient conditions avoid the 
construction of redundant interpretations~\cite{Console96}. Another strategy 
entails additional restrictions in the amount of computer memory and time needed 
to run the algorithm, resulting in a selective pruning of the node expansion 
while sacrificing optimality; this strategy is similar to the one used in the 
K-Beam algorithm~\cite{Edelkamp2011}. 

The \begin{small}CONSTRUE()\end{small} algorithm is based on the assumption that 
all the evidence to be explained is available at the beginning of the 
interpretation task. A new version of the algorithm should be provided to cope 
with a wide range of problems, where the interpretation must be updated as new 
evidence becomes available over time. Examples of such problems are continuous 
biosignal monitoring or plan execution monitoring~\cite{Bartak14}. At the 
emergence of a new piece of evidence, two reasoning modes may come into play 
triggered by the \begin{small}CONSTRUE()\end{small} algorithm: a new explanatory 
hypothesis can be conjectured by means of the \begin{small}ABDUCE()\end{small} 
procedure, or the evidence can be incorporated in an existing hypothesis by 
means of the \begin{small}SUBSUME()\end{small} procedure. In this way, the 
incorporation of new evidence over time is seamlessly integrated into the 
hypothesize-and-test cycle. Furthermore, to properly address these 
interpretation scenarios, the heuristics used to guide the search must be 
updated to account for the timing of the interpretation process, which will lead 
to the definition of a covering ratio until time $t$, and a complexity until 
time $t$.

\section*{Implementation}
\addcontentsline{toc}{section}{\protect\numberline{}Implementation}%

With the aim of supporting reproducible research, the full source code of the 
algorithms presented in this paper has been published under an Open Source 
License\footnote{\url{https://github.com/citiususc/construe}}, along with a 
knowledge base for the interpretation of the ECG signal strips of all examples 
in this paper.

\section*{Acknowledgments}
\addcontentsline{toc}{section}{\protect\numberline{}Acknowledgements}%

This work was supported by the Spanish Ministry of Economy and Competitiveness
under project TIN2014-55183-R. T. Teijeiro was funded by an FPU grant from the
Spanish Ministry of Education (MEC) (ref. AP2010-1012).

\section*{References}
\addcontentsline{toc}{section}{\protect\numberline{}References}%

\bibliographystyle{plain}
\bibliography{bibliography}

\begin{thebibliography}{10}

\bibitem{Aho06}
A.V. Aho, M.S. Lam, R.~Sethi, and J.D. Ullman.
\newblock {\em {Compilers: Principles, Techniques and Tools}}.
\newblock Pearson Education, Inc., 2006.

\bibitem{Barro94a}
S.~Barro, R.~Mar{\'i}n, J.~Mira, and A.~Pat{\'o}n.
\newblock {A model and a language for the fuzzy representation and handling of
  time}.
\newblock {\em Fuzzy Sets and Systems}, 61:153--175, 1994.

\bibitem{Bartak14}
R.~Bart{\'a}k, R.~A. Morris, and K.~B. Venable.
\newblock {An Introduction to Constraint-Based Temporal Reasoning}.
\newblock {\em Synthesis Lectures on Artificial Intelligence and Machine
  Learning}, 8(1):1--121, feb 2014.

\bibitem{Brusoni98}
V.~Brusoni, L.~Console, P.~Terenziani, and D.~Theseider Dupr{\'e}.
\newblock {A spectrum of definitions for temporal model-based diagnosis}.
\newblock {\em Artificial Intelligence}, 102(1):39--79, 1998.

\bibitem{Chakravarty00}
S.~Chakravarty and Y.~Shahar.
\newblock {CAPSUL: A constraint-based specification of repeating patterns in
  time-oriented data}.
\newblock {\em Annals of Mathematics and Artificial Intelligence}, 30:3--22,
  2000.

\bibitem{Charniak88}
E.~Charniak.
\newblock {Motivation analysis, abductive unification and nonmonotonic
  equality}.
\newblock {\em Artificial Intelligence}, 34(3):275--295, 1989.

\bibitem{Chow70}
C.K. Chow.
\newblock {On optimum recognition error and reject tradeoff}.
\newblock {\em IEEE Transaction on Information Theory}, 16(1):41--46, 1970.

\bibitem{Clifford17}
G.~Clifford, C.~Liu, B.~Moody, I.~Silva, Q.~Li, A.~Johnson, and R.~Mark.
\newblock {AF Classification from a Short Single Lead ECG Recording: the
  PhysioNet Computing in Cardiology Challenge}.
\newblock In {\em {Proceedings of the 2017 Computing in Cardiology Conference
  (CinC)}}, volume~47, 2017.

\bibitem{Console96}
L.~Console, L.~Portinale, and D.~Theseider Dupr{\'e}.
\newblock {Using compiled knowledge to guide and focus abductive diagnosis}.
\newblock {\em IEEE Transactions on Knowledge and Data Engineering},
  8(5):690--706, 1996.

\bibitem{Console91a}
L.~Console and P.~Torasso.
\newblock {A spectrum of logical definitions of model-based diagnosis}.
\newblock {\em Computational Intelligence}, 3(7):133--141, 1991.

\bibitem{CSE85}
Working~Party CSE.
\newblock {Recommendations for measurement standards in quantitative
  electrocardiography.}
\newblock {\em European Heart Journal}, 6(10):815--825, 1985.

\bibitem{Dechter03}
R.~Dechter.
\newblock {\em {Constraint Processing}}.
\newblock Morgan Kaufmann Publishers, 2003.

\bibitem{Dubois95}
D.~Dubois and H.~Prade.
\newblock {Fuzzy relation equations and causal reasoning}.
\newblock {\em Special Issue on ''Equations and Relations on Ordered Structures
  : Mathematical Aspects and Applications" (A. Di Nola, W. Pedrycz, S. Sessa,
  eds.), Fuzzy Sets and Systems}, 75:119--134, 1995.

\bibitem{Edelkamp2011}
S.~Edelkamp and S.~Schr{\"o}dl.
\newblock {\em {Heuristic Search: Theory and Applications}}.
\newblock Morgan Kaufmann, 2011.

\bibitem{Ferrucci2012}
D.~Ferrucci, A.~Levas, S.~Bagchi, D.~Gondek, and E.T. Mueller.
\newblock {Watson: Beyond Jeopardy}.
\newblock {\em Artificial Intelligence}, 199--200:93--105, 2012.

\bibitem{Flach96}
P.~Flach.
\newblock {Abduction and induction: Syllogistic and inferential perspectives}.
\newblock In {\em {Abductive and Inductive Reasoning Workshop Notes}}, pages
  31--35. University of Bristol, 1996.

\bibitem{Fumera00}
G.~Fumera, F.~Roli, and G.~Giacinto.
\newblock {Reject option with multiple thresholds}.
\newblock {\em Pattern Recognition}, (33):2099--2101, 2000.

\bibitem{Goldberger00}
A.~L. Goldberger et~al.
\newblock {PhysioBank, PhysioToolkit, and PhysioNet: Components of a New
  Research Resource for Complex Physiologic Signals}.
\newblock {\em Circulation}, 101(23):215--220, June 2000.

\bibitem{Haimowitz93}
I.J. Haimowitz and I.S. Kohane.
\newblock {Automated Trend Detection with Alternate Temporal Hypotheses}.
\newblock In {\em {Proceedings of the 13th International Joint Conference of
  Artificial Intelligence}}, volume~1, pages 146--151, 1993.

\bibitem{Haimowitz95}
I.J. Haimowitz, P.P. Le, and I.S. Kohane.
\newblock {Clinical monitoring using regression-based trend templates}.
\newblock {\em Artificial Intelligence in Medicine}, 7(6):473--496, 1995.

\bibitem{Peirce31}
C.~Hartshorn et~al.
\newblock {\em {Collected papers of Charles Sanders Peirce}}.
\newblock Harvard University Press, 1931.

\bibitem{Hobbs93}
J.R. Hobbs, M.~Stickel, and P.~Martin.
\newblock {Interpretation as abduction}.
\newblock {\em Artificial Intelligence}, 63:69--142, 1993.

\bibitem{Hopcroft01}
J.~Hopcroft, R.~Motwani, and J.~Ullman.
\newblock {\em {Introduction to automata theory, languages and computation}}.
\newblock Addison-Wesley, 2001.

\bibitem{Josephson94}
J.R. Josephson and S.G. Josephson.
\newblock {\em {Abductive inference. Computation, philosophy, technology}}.
\newblock Cambridge University Press, 1994.

\bibitem{Juarez08}
J.~M. Ju{\'a}rez, M.~Campos, J.~Palma, and R.~Mar{\'i}n.
\newblock {Computing context-dependent temporal diagnosis in complex domains}.
\newblock {\em Expert Systems with Applications}, 35(3):991--1010, 2008.

\bibitem{Laguna94}
P.~Laguna, R.~Jan{\'e}, and P.~Caminal.
\newblock {Automatic detection of wave boundaries in multilead ECG signals:
  validation with the CSE database}.
\newblock {\em Computers and Biomedical Research}, 27:45--60, 1994.

\bibitem{Larizza95}
C.~Larizza, G.~Bernuzzi, and M.~Stefanelli.
\newblock {A general framework for building patient monitoring systems}.
\newblock In {\em {Proceedings of the 5th Conference on Artificial intelligence
  in Medicine}}, pages 91--102, 1995.

\bibitem{Litman87}
D.~Litman and J.~Allen.
\newblock {A plan recognition model for subdialogues in conversation}.
\newblock {\em {Cognitive Science}}, 11:163--200, 1987.

\bibitem{Luz15}
E.~J.~S. Luz, W.~R. Schwartz, G.~C{\'a}mara-Ch{\'a}vez, and D.~Menotti.
\newblock {ECG-based Heartbeat Classification for Arrhythmia Detection: A
  Survey}.
\newblock {\em Computer Methods and Programs in Biomedicine}, 2016.

\bibitem{Morchen03}
F.~M{\"o}rchen.
\newblock {Time series feature extraction for data mining using DWT and DFT}.
\newblock Technical Report no. 33, Department of Mathematics and Computer
  Science, University of Marburg, 2003.

\bibitem{Nauck99}
D.~Nauck and R.~Kruse.
\newblock {Obtaining interpretable fuzzy classification rules from medical
  data}.
\newblock {\em Artificial Intelligence in Medicine}, 16(2):149--169, 1999.

\bibitem{Palma06}
J.~Palma, J.~M. Ju{\'a}rez, M.~Campos, and R.~Mar{\'i}n.
\newblock {Fuzzy theory approach for temporal model-based diagnosis: An
  application to medical domains}.
\newblock {\em Artificial Intelligence in Medicine}, 38(2):197, 2006.

\bibitem{Peng90}
Y.~Peng and J.A. Reggia.
\newblock {\em {Abductive inference models for diagnostic problem-solving}}.
\newblock Springer-Verlag, 1990.

\bibitem{Petrenas15}
A.~Petrenas, V.~Marozas, and L.~S{\"o}rnmo.
\newblock {Low-complexity detection of atrial fibrillation in continuous
  long-term monitoring.}
\newblock {\em Computers in biology and medicine}, 65:184--91, oct 2015.

\bibitem{Pimentel14}
M.A.F. Pimentel, D.A. Clifton, L.~Clifton, and L.~Tarassenko.
\newblock {A review of novelty detection}.
\newblock {\em Signal Processing}, (99):215--249, 2014.

\bibitem{Poole90}
D.~Poole.
\newblock {A methodology for using a default and abductive reasoning system}.
\newblock {\em International Journal of Intelligent Systems}, 5(5):521--548,
  1990.

\bibitem{Poole00}
D.~Poole.
\newblock {Learning, Bayesian Probability, Graphical Models, and Abduction}.
\newblock In {\em {Abduction and Induction: Essays on their Relation and
  Integration}}, pages 153--168. Springer Netherlands, 2000.

\bibitem{Sacchi15}
L.~Sacchi, E.~Parimbelli, S.~Panzarasa, N.~Viani, E.~Rizzo, C.~Napolitano,
  R.~Ioana Budasu, and S.~Quaglini.
\newblock {Combining Decision Support System-Generated Recommendations with
  Interactive Guideline Visualization for Better Informed Decisions}.
\newblock In {\em {Artificial Intelligence in Medicine}}, pages 337--341.
  Springer International Publishing, 2015.

\bibitem{Shahar97}
Y.~Shahar.
\newblock {A framework for knowledge-based temporal abstraction}.
\newblock {\em Artificial intelligence}, 90(1--2):79--133, 1997.

\bibitem{Shahar98}
Y.~Shahar.
\newblock {Dynamic temporal interpretation contexts for temporal abstraction}.
\newblock {\em Annals of Mathematics and Artificial Intelligence},
  22(1--2):159--192, 1998.

\bibitem{Shahar99}
Y.~Shahar.
\newblock {Knowledge-based temporal interpolation}.
\newblock {\em Journal of experimental and theoretical artificial
  intelligence}, 11:123--144, 1999.

\bibitem{Shahar96}
Y.~Shahar and M.A. Musen.
\newblock {Knowledge-based temporal abstraction in clinical domains}.
\newblock {\em Artificial Intelligence in Medicine}, 8(3):267--298, 1996.

\bibitem{Teijeiro14}
T.~Teijeiro, P.~F{\'e}lix, and J.~Presedo.
\newblock {Using Temporal Abduction for Biosignal Interpretation: A Case Study
  on QRS Detection}.
\newblock In {\em {2014 IEEE International Conference on Healthcare
  Informatics}}, pages 334--339, 2014.

\bibitem{Teijeiro16}
T.~Teijeiro, P.~F{\'e}lix, J.~Presedo, and D.~Castro.
\newblock {Heartbeat classification using abstract features from the abductive
  interpretation of the ECG}.
\newblock {\em IEEE Journal of Biomedical and Health Informatics}, 2016.

\bibitem{Teijeiro17}
T.~Teijeiro, C.A. Garc{\'i}a, D.~Castro, and P.~F{\'e}lix.
\newblock {Arrhythmia Classification from the Abductive Interpretation of Short
  Single-Lead ECG Records}.
\newblock In {\em {Proceedings of the 2017 Computing in Cardiology Conference
  (CinC)}}, volume~47, 2017.

\bibitem{Marriott08}
Galen~S. Wagner.
\newblock {\em {Marriott's Practical Electrocardiography}}.
\newblock Wolters Kluwer Health/Lippincott Williams \& Wilkins, 11 edition,
  2008.

\bibitem{Zong03}
W.~Zong, G.B. Moody, and D.~Jiang.
\newblock {A robust open-source algorithm to detect onset and duration of QRS
  complexes}.
\newblock In {\em {Computers in Cardiology}}, pages 737--740, 2003.

\end{thebibliography}

\end{document}